\documentclass{article}

\usepackage[preprint]{neurips_2026}

\usepackage[utf8]{inputenc}
\usepackage[T1]{fontenc}
\usepackage{microtype}

\usepackage{amsmath}
\usepackage{amsfonts}
\usepackage{nicefrac}

\usepackage{graphicx}
\usepackage{float}
\usepackage{multirow}
\usepackage{booktabs}
\usepackage{subcaption}
\usepackage{pgfplots}
\pgfplotsset{compat=1.18}

\usepackage{enumitem}

\usepackage{listings}
\usepackage[most,breakable]{tcolorbox}
\usepackage{xcolor}

\usepackage{url}
\usepackage{hyperref}
\hypersetup{
  hidelinks,
  pdftitle={Better Reasoning, Not Just Bigger Models: A Structure-Grounded Framework for Efficient and Reliable Inference},
  pdfauthor={Liangyan Li, Amirhossein Aghaei, Shiyi Yang, Shaohua Wang, Yan Wang}
}

\title{
  Better Reasoning, Not Just Bigger Models:
  A Structure-Grounded Framework for
  Efficient and Reliable Inference
}

\author{
  Amirhossein Aghaei\thanks{Equal contribution.} \\
  Huawei Research Canada \\
  Markham, Canada \\
  \texttt{amirhossein.aghaei@h-partners.com}
  \And
  Liangyan Li\footnotemark[1] \\
  Huawei Research Canada \\
  Waterloo, Canada \\
  \texttt{lillian.li2@h-partners.com}
  \And
  Shiyi Yang \\
  Huawei Research Canada \\
  Waterloo, Canada \\
  \texttt{shiyi.yang1@huawei.com}
  \AND
  Shaohua Wang \\
  Huawei Technologies Co., Ltd. \\
  Beijing, China \\
  \texttt{wangshaohua25@huawei.com}
  \And
  Yan Wang \\
  Central University of Finance and Economics \\
  Beijing, China \\
  \texttt{dayanking@gamil.com}
}

\date{}

\begin{document}

\maketitle
\begin{abstract}
Scaling large language models has substantially improved linguistic fluency, yet reliable autonomous agents require more than fluent generation. In multi-turn inference, current agentic systems often produce semantically plausible but structurally ungrounded reasoning, especially under incomplete observations. We propose \textbf{Structure-Grounded Uncertainty Reduction (SGUR)}, a training-free framework that complements token-level generation with active uncertainty reduction over structured hypothesis spaces. SGUR represents incomplete observations as a query graph and aligns them with candidate hypothesis graphs through a prior-weighted partial graph matching mechanism, where database-derived empirical priors assign greater importance to decision-relevant relations rather than treating all graph edges uniformly. The resulting inference state is iteratively refined by an active information-gain strategy that selects the next most informative node to query, systematically pruning the candidate space. We instantiate SGUR on a challenging PrimeKG-derived sequential medical diagnosis benchmark, demonstrating that structural grounding can partially substitute for model scale: starting from only two observed symptoms, SGUR achieves \textbf{92.5\% Hit@3}, outperforming the strongest Graph-RAG baseline by \textbf{9.4 percentage points} while using approximately \textbf{$5\times$ fewer tokens}. Notably, SGUR enables smaller backbone models to match or exceed the performance of models up to \textbf{$10\times$ larger}, suggesting that structural grounding provides a more robust and efficient paradigm for scaling agent performance than relying exclusively on increasing model size.

\end{abstract}

\begin{figure}[t]
    \centering
    \includegraphics[width=0.9\linewidth]{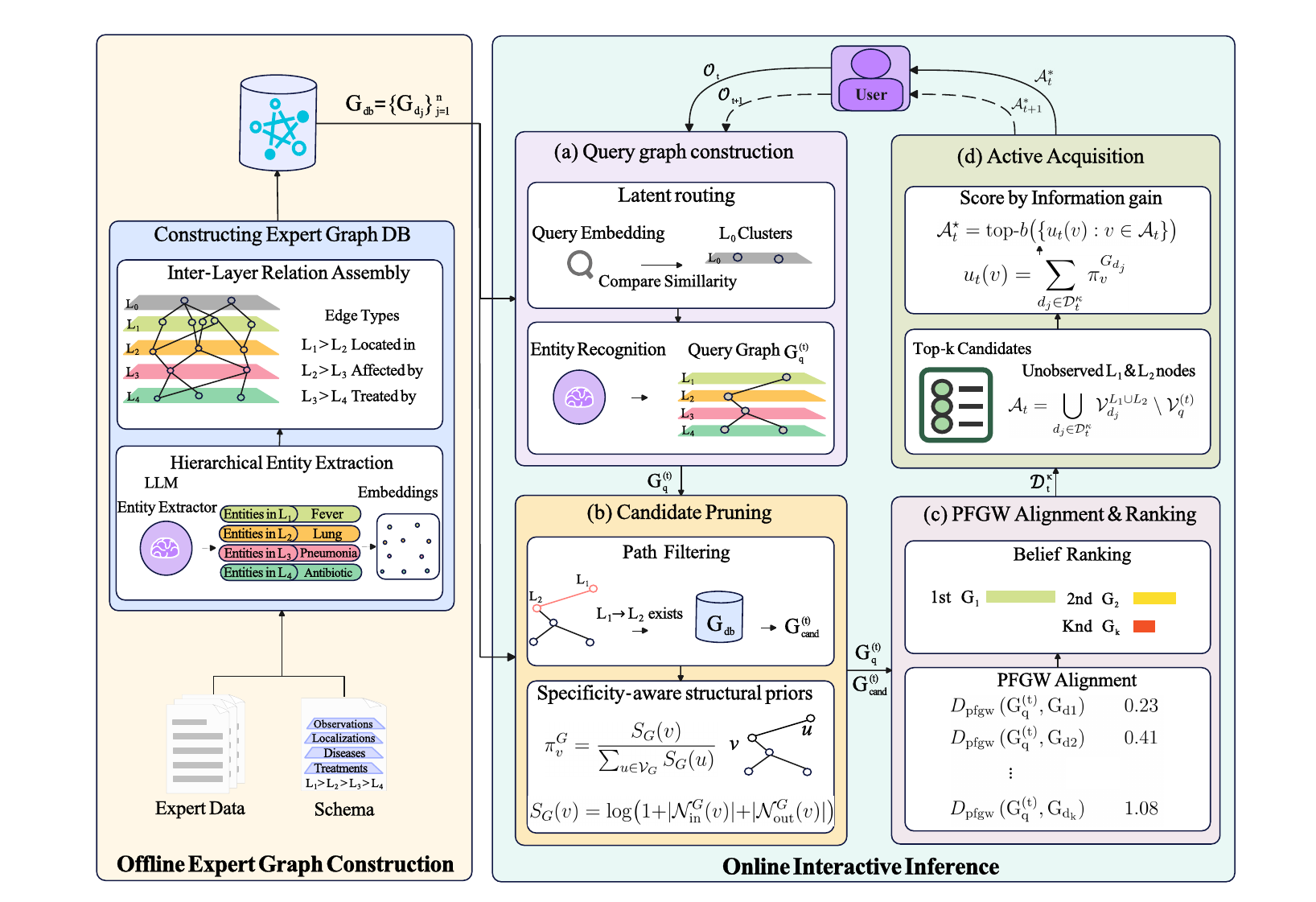}
    \caption{Overview of the SGUR framework. The offline phase constructs a hierarchical expert graph database \(G_{\mathrm{db}}\) from expert data under a shared schema. The online phase performs closed-loop multi-turn inference through four stages: (a) query graph construction, which grounds observations \(\mathcal{O}_t\) into \(G_q^{(t)}\) with latent routing; (b) candidate pruning, which applies path filtering and specificity-aware structural priors to obtain \({G}_{\mathrm{cand}}^{(t)}\); (c) prior-weighted partial FGW alignment, which ranks candidate graphs by structural consistency with the query; and (d) active acquisition, which selects informative unobserved nodes to obtain \(\mathcal{O}_{t+1}\). Across rounds, SGUR progressively contracts the candidate space under partial observation.}
    \label{fig:pipeline}
\end{figure}

\section{Introduction}
The design of autonomous agents capable of reasoning under \emph{partial, noisy, and sequentially revealed information} remains a central challenge for building reliable and efficient agentic systems. Although scaling large language models (LLMs) has substantially improved linguistic fluency\citep{openai_embeddings_2024,openai_gpt52_2025,openai_gpt52_systemcard_2025}, reliable multi-turn inference requires more than fluent generation. In complex decision-making environments, an agent must maintain a coherent hypothesis space, strategically acquire missing evidence, and update its belief state as new information arrives. However, many current agentic systems still rely primarily on token-level generation and linguistic similarity, which can lead to weakly grounded multi-turn inference when semantic evidence is not jointly constrained by relational structure\citep{fastowski2024understanding,zheng2025rethinking}. Under iterative interaction, such systems may exhibit \emph{heuristic substitution}\citep{kahneman2011thinking}, relying on associative pattern matching rather than refining hypotheses against jointly grounded semantic evidence and relational structure.

Existing paradigms only partially address this challenge. Chain-of-Thought and Tree-of-Thought improve reasoning decomposition, but continue to operate largely within the model's internal probabilistic space\citep{wei2022chain,yao2023tree}. Recent memory-augmented systems also compress conversational history, yet topology-insensitive compression can discard decision-relevant cues\citep{you2026d}. GraphRAG frameworks introduce additional structure by retrieving graph-organized evidence\citep{edge2024local,guo2024lightrag,peng2025graph,zhang2025survey}. Recent hierarchical GraphRAG methods further strengthen this line by organizing knowledge into attributed communities and hierarchical indexes, improving retrieval relevance and reducing online token cost\citep{wang2026archrag}. However, their main contribution remains on the retrieval side. The graph is primarily used to access supporting evidence, which is then flattened into textual context for downstream generation. Even with hierarchical indices such as ArchRAG\citep{wang2026archrag}, the reasoning process can remain \textbf{structurally underconstrained}. Retrieved communities are typically converted into textual context for generation, while graph topology and relational distances do not directly define the inference objective. This limits the agent's ability to use the knowledge graph as an active hypothesis space rather than only as a retrieval substrate.

This limitation is especially consequential in high-stakes sequential settings, where deployment-relevant agent performance depends on uncovering missing evidence through iterative, goal-directed interaction\citep{bobrova2026clinical}. Sequential, multi-hop clinical diagnosis provides a representative example, where each intermediate inference can shape subsequent evidence acquisition and hypothesis revision, a process recently formalized as the \emph{Sequential Diagnosis} paradigm\citep{nori2025sequential,trivedi2023ircot}. As a result, early reasoning errors may propagate across turns and compromise the final decision\citep{xinjie2025reagent}. Recent empirical analyses further support this concern. LLMs often become path-dependent in extended interactions, committing early to brittle assumptions and failing to recover once the dialogue drifts\citep{laban2025llms}. Moreover, from an information-theoretic perspective, multi-turn interaction can exhibit diminishing information gain and increasing redundancy across turns\citep{gorle2025quantifying}. Although recent methods introduce iterative exploration, their step-wise search often remains heuristic, making them vulnerable to shallow retrieval, error accumulation, and task-shift failures\citep{luo2023reasoning,sun2023think,lv2026all}. These challenges motivate a shift from likelihood-driven generation and retrieval-augmented context construction toward semantic-structural uncertainty reduction over organized hypothesis spaces\citep{scanlin2026representation}.

In this paper, we introduce \textbf{S}tructure-\textbf{G}rounded \textbf{U}ncertainty \textbf{R}eduction (\textbf{SGUR}), a training-free framework for efficient and reliable multi-turn inference. SGUR reformulates \textbf{multi-step reasoning as uncertainty reduction over structured hypothesis spaces}, rather than unconstrained free-form generation. 
At its core, SGUR couples geometric structural alignment with probabilistic belief updating. It maps the alignment cost between fragmented observations and candidate knowledge graphs into a likelihood score, enabling a principled update of the agent's belief state. By jointly grounding each inference step in semantic evidence, relational structure, and domain-specific priors, SGUR constrains the evolving reasoning trajectory to remain both semantically aligned and structurally coherent under incomplete observations.

Operationally, SGUR uses two coupled mechanisms. (i) a \textbf{prior-weighted partial alignment} that enables robust semantic-structural matching under sparse, incomplete evidence without requiring full graph correspondence; and (ii) an \textbf{Expected Information Gain (EIG)-driven acquisition strategy} that selects the next most informative query node. This cycle ensures that both hypothesis refinement and evidence acquisition are governed by shared semantic-structural constraints, leading to a rapid and efficient collapse of the candidate hypothesis space.

We evaluate SGUR in sequential clinical diagnosis over a large-scale hierarchical biomedical knowledge graph, a demanding setting in which evidence is incomplete, observations are revealed iteratively, and reliability is essential. Unlike small single-turn question-answering benchmarks, this setting requires the agent to reason over a structured hypothesis space, acquire missing evidence across turns, and maintain consistency over long-horizon inference. Empirically, SGUR delivers significant system-level gains in diagnostic accuracy, inference efficiency, and token cost across strong baselines. Crucially, when paired with smaller backbone models, SGUR achieves performance comparable to or better than baselines powered by substantially larger state-of-the-art models. These results show that reliable agent performance depends not only on model scale, but also on \textbf{the quality of the reasoning framework that organizes inference}.

In summary, our contributions are threefold. First, we introduce a training-free \textbf{inference paradigm} that reformulates multi-turn reasoning as Bayesian uncertainty contraction over structured hypothesis spaces, providing a principled alternative to unconstrained free-form generation. Second, we develop a unified \textbf{semantic-structural reasoning mechanism} that couples prior-weighted partial graph alignment with information-theoretic evidence acquisition, enabling robust belief refinement and goal-directed probing under sparse and sequentially revealed observations. Third, we provide a large-scale evaluation on a hierarchy-structured sequential diagnosis benchmark, demonstrating an empirical \textbf{structure-scale substitution effect}: SGUR improves Hit@3 by 9.4 percentage points over the strongest GraphRAG baseline while reducing token overhead by approximately $5\times$, and enables smaller backbone models to match or exceed substantially larger LLMs.

\section{Related Work}
\label{sec:related}
\paragraph{Reasoning and Retrieval under Weak Structural Control.}
Recent advances in LLM reasoning have improved performance through structured prompting, retrieval augmentation, and graph-organized evidence. Structured prompting methods, including CoT, ToT, GoT, least-to-most prompting, and self-consistency, improve decomposition and deliberative search\citep{wei2022chain, yao2023tree, besta2024graph, zhou2022least, wang2022self}. In parallel, RAG-style systems improve factual grounding by incorporating external evidence into generation\citep{lewis2020retrieval, gao2023retrieval, borgeaud2022improving}. Hybrid approaches such as RAT and BayesRAG variants further interleave retrieval with reasoning or verification\citep{wang2024rat, li2026bayesrag, yang2026autothinkrag, wan2025mmgraphrag, jiang2023active}. GraphRAG methods extend retrieval-augmented reasoning by organizing evidence into explicit graph structures for local and global querying\citep{edge2024local, guo2024lightrag, hu2025grag, zhu2025knowledge}, including medical variants such as MedGraphRAG and KARE\citep{wu2025medical, jiang2024reasoning}. Recent hierarchical GraphRAG variants further improve retrieval relevance and efficiency through community-level organization and hierarchical indexing\citep{wang2026archrag}.

However, these methods still organize reasoning primarily in prompt, text, or retrieval space. Even when graphs are introduced, they are typically used as retrieval scaffolds whose outputs are flattened into textual context for downstream generation\citep{edge2024local, wu2025medical, li2024structrag, gutierrez2024hipporag, wang2026archrag}. As a result, they may improve evidence access without explicitly constraining the evolving reasoning trajectory itself\citep{guo2024lightrag, li2024structrag, chen2024multi, liang2025kag, han2025rag, xiang2025use}. The central gap is therefore not only how to retrieve better evidence, but how to make graph structure part of the inference state itself.

\paragraph{Geometric Alignment and Active Uncertainty Reduction.} Optimal Transport (OT) provides a principled foundation for graph alignment under soft correspondence. Gromov--Wasserstein (GW) and Fused Gromov--Wasserstein (FGW) compare structured objects by jointly accounting for node semantics and relational structure\citep{memoli2011gromov, titouan2019optimal, vayer2020fused, vayer2018optimal}, while scalable approximations improve practicality for larger graphs\citep{ scetbon2022linear, nguyen2023linear}. Partial transport methods relax the full-correspondence requirement and instead focus alignment on overlapping substructures, making them especially relevant under incomplete observations\citep{chapel2020partial, bai2025fused}.

A complementary line of work treats reasoning as an interactive process of sequential uncertainty reduction rather than one-shot generation\citep{yao2023react, trivedi2023ircot}. In medical settings, this appears in active question selection and information-seeking diagnosis\citep{hu2024uot, wong2026medclarify, wang2024healthq, nori2025sequential}, while related systems explore adaptive inference control, solver arbitration, and memory management\citep{bergamaschi2025fast,zhang2025leveraging, you2026d, chung2025thinker, hu2026beyond}. Yet these approaches generally manage alignment, questioning, and belief refinement as separate procedures in text, memory, or retrieval space. SGUR differs by coupling semantic-structural partial alignment with EIG-driven evidence acquisition within a shared belief state, so that both hypothesis refinement and question selection are governed by the same structure-grounded inference objective.

\section{Methodology: The SGUR Framework}
\label{sec:methodology}
SGUR follows the two-phase framework in Figure~\ref{fig:pipeline}. 
Section~\ref{sec:method:offline} describes the offline phase on the left, where expert data are organized into a fixed hierarchical graph database \(G_{\mathrm{db}}\) under a shared diagnostic schema. 
Section~\ref{sec:method:online} describes the online phase on the right, where each interaction round follows four figure-aligned stages: 
(a) latent routing first aligns the current observations \(\mathcal{O}_t\) with database-level observation anchors and uses the matched anchors to construct the query graph \(G_q^{(t)}\); 
(b) candidate pruning retrieves the associated candidate hypothesis graphs \({G}_{\mathrm{cand}}^{(t)}\) using path filtering and specificity-aware structural priors; 
(c) prior-weighted partial Fused Gromov--Wasserstein (PFGW) alignment compares \(G_q^{(t)}\) with each candidate graph and induces a belief ranking; and 
(d) active acquisition selects informative unobserved nodes to obtain \(\mathcal{O}_{t+1}\). 
Together, the fixed expert database and the online loop define SGUR as a structure-grounded inference procedure that progressively contracts the candidate hypothesis space under partial observation.
\subsection{Offline Expert Graph Construction}
\label{sec:method:offline}

The offline phase constructs the fixed expert graph database used by the online inference loop. As shown on the left side of Figure~\ref{fig:pipeline}, SGUR extracts hierarchical entities from expert data, maps them into a shared diagnostic schema, and assembles directed inter-layer relations between adjacent semantic layers. In our PrimeKG-derived clinical instantiation, the schema contains observations (\(L_1\)), localizations (\(L_2\)), candidate diagnoses (\(L_3\)), and treatments (\(L_4\)). SGUR also builds an auxiliary \(L_0\) routing layer by clustering observation embeddings; \(L_0\) serves as a routing index for fast query-graph construction, not as a diagnostic semantic layer.

Formally, the expert graph database is
\begin{equation}
{G}_{\mathrm{db}}=\{G_{d_j}\}_{j=1}^{n},
\end{equation}
where each diagnosis-centered graph is
\begin{equation}
G_{d_j}=(\mathcal{V}_{d_j},\mathcal{E}_{d_j},\omega_{d_j},\mathbf{X}_{d_j},\ell).
\end{equation}
Here, \(\mathcal{V}_{d_j}\) is the node set, \(\mathcal{E}_{d_j}\subseteq \mathcal{V}_{d_j}\times\mathcal{V}_{d_j}\) is the directed edge set, \(\mathbf{X}_{d_j}\) stores node features, and \(\ell:\mathcal{V}_{d_j}\rightarrow\{L_1,L_2,L_3,L_4\}\) maps nodes to semantic layers. In our benchmark, each \(d_j\) corresponds to a disease, but the construction only requires a structured diagnostic schema.

The edge weights \(\omega_{d_j}\) instantiate the specificity-aware structural priors. For each outgoing relation \((u,v)\in\mathcal{E}_{d_j}\), SGUR uses
\begin{equation}
\omega_{uv}=\frac{1}{|\mathcal{N}_{\mathrm{out}}(u)|},
\label{eq:edge_weight}
\end{equation}
where \(\mathcal{N}_{\mathrm{out}}(u)\) denotes the outgoing neighbors of \(u\). This down-weights generic nodes with many downstream continuations and emphasizes more discriminative relations during candidate pruning and prior-weighted partial FGW alignment. During online inference, \({G}_{\mathrm{db}}\) is kept fixed: SGUR grounds current observations, routes them to a candidate subset, aligns the query graph with candidate graphs, and updates the belief ranking without reconstructing or updating the expert database.

\subsection{Online Interactive Inference}
\label{sec:method:online}

The online phase performs closed-loop inference over the fixed database \({G}_{\mathrm{db}}\). At round \(t\), the accumulated evidence \(\mathcal{O}_t\) is represented as a partial query graph
\begin{equation}
G_q^{(t)}=\Gamma(\mathcal{O}_t)
=(\mathcal{V}_q^{(t)},\mathcal{E}_q^{(t)},\omega_q^{(t)},\mathbf{X}_q^{(t)},\ell_q^{(t)}),
\end{equation}
where \(\Gamma(\cdot)\) maps extracted entities into the shared schema. The online loop refines a belief ranking over candidate graphs while acquiring the next evidence that is most useful under the current ranking.

\paragraph{(a) Query graph construction.}
SGUR first grounds the current observations into \(G_q^{(t)}\). To avoid aligning every query against the full database, the auxiliary \(L_0\) routing layer localizes the query to a coarse observation neighborhood before fine-grained matching. After passing through \(L_0\), entities are extracted from the query (see Appendix~\ref{app:prompts} for the entity recognition prompt) and used to construct the query graph \(G_q^{(t)}\). This corresponds to the query graph construction block in Figure~\ref{fig:pipeline}.

\paragraph{(b) Candidate pruning and structural priors.}
After latent routing, SGUR applies path filtering to retain only candidate graphs that are structurally compatible with the current query:
\begin{equation}
{G}_{\mathrm{cand}}^{(t)}
=
\left\{
G_{d_j}\in{G}_{\mathrm{db}}
:
R_0(G_q^{(t)},G_{d_j})=1,\;
F_{\mathrm{path}}(G_q^{(t)},G_{d_j})=1
\right\},
\end{equation}
where \(R_0\) denotes the \(L_0\)-based routing condition and \(F_{\mathrm{path}}\) denotes inter-layer path compatibility. SGUR reuses the specificity-aware relation weights in Eq.~\eqref{eq:edge_weight}. For node-level saliency, for any graph \(G\) we define
\begin{equation}
S_G(v)=\log\!\big(1+|\mathcal{N}^{G}_{\mathrm{in}}(v)|+|\mathcal{N}^{G}_{\mathrm{out}}(v)|\big),
\qquad
\pi_v^{G}=\frac{S_G(v)}{\sum_{u\in\mathcal{V}_{G}}S_G(u)}.
\label{eq:node_saliency}
\end{equation}
The normalized saliency \(\pi_v^G\) provides node masses for alignment and utility scores for active acquisition.

\paragraph{(c) Prior-weighted partial FGW alignment and belief ranking.}
For each \(G_{d_j}\in{G}_{\mathrm{cand}}^{(t)}\), SGUR compares \(G_q^{(t)}\) with \(G_{d_j}\) using prior-weighted PFGW alignment. Let \(\mathbf{M}^{(j,t)}\) be the node feature discrepancy matrix, instantiated as cosine distances between query-node and candidate-node embeddings. Let \(C_q^{(t)}\) and \(C_{d_j}\) denote weighted shortest-path distance matrices encoding internal graph structure. We set the candidate and query node masses as
\(\mathbf{p}^{(j)}=(\pi_a^{G_{d_j}})_{a\in\mathcal{V}_{d_j}}\) and
\(\mathbf{q}^{(t)}=(\pi_b^{G_q^{(t)}})_{b\in\mathcal{V}_q^{(t)}}\).
SGUR computes
\begin{equation}
\begin{aligned}
D_{\mathrm{PFGW}}(G_q^{(t)},G_{d_j})
&= \min_{\gamma\in\Pi_m(\mathbf{p}^{(j)},\mathbf{q}^{(t)})}
\Bigl\{(1-\alpha)\langle\gamma,\mathbf{M}^{(j,t)}\rangle \\
&\qquad {}+ \alpha\!\!\sum_{a,a',b,b'}\!
\left|C_{d_j}(a,a')-C_q^{(t)}(b,b')\right|^2
\gamma_{ab}\gamma_{a'b'}\Bigr\}.
\end{aligned}
\label{eq:pfgw}
\end{equation}
where \(\Pi_m(\mathbf{p}^{(j)},\mathbf{q}^{(t)})\) is the set of partial couplings with transported mass \(m\). The partial coupling allows unmatched query evidence and unmatched expert structure to remain uncoupled, which is essential when the query covers only a sparse substructure of the candidate graph.

The resulting discrepancies are converted into a normalized belief ranking:
\begin{equation}
P^{(t)}(d_j)
=
\frac{\exp(-D_{\mathrm{PFGW}}(G_q^{(t)},G_{d_j})/\beta)}
{\sum_{G_{d_k}\in{G}_{\mathrm{cand}}^{(t)}}\exp(-D_{\mathrm{PFGW}}(G_q^{(t)},G_{d_k})/\beta)},
\label{eq:belief}
\end{equation}
where \(\beta>0\) controls ranking sharpness. The induced rank is
\begin{equation}
r_t(d_j)=\mathrm{rank}\!\big(P^{(t)}(d_j)\big),
\qquad
G_{d_j}\in{G}_{\mathrm{cand}}^{(t)}.
\end{equation}
We measure remaining uncertainty by the entropy of this candidate belief distribution:
\begin{equation}
H^{(t)}
=
-\sum_{G_{d_j}\in{G}_{\mathrm{cand}}^{(t)}}
P^{(t)}(d_j)\log P^{(t)}(d_j).
\end{equation}
Thus, candidate-space contraction is realized as progressive concentration of \(P^{(t)}\) across interaction rounds.

\paragraph{(d) Active acquisition.}
SGUR selects the next evidence from unobserved \(L_1/L_2\) nodes in the top-\(\kappa\) candidate graphs. The ideal objective is to choose the query that most reduces uncertainty in the belief distribution:
\begin{equation}
\mathrm{EIG}(s)
=
\mathbb{E}_{r\sim p(r\mid s,P^{(t)})}
\big[
H^{(t)}-H(P^{(t+1)}\mid r,s)
\big].
\label{eq:eig}
\end{equation}
Exact EIG requires a response model over all possible user answers. SGUR therefore uses the structure-guided proxy shown in Figure~\ref{fig:pipeline}. Let \(\mathcal{D}_t^\kappa\) be the top-\(\kappa\) candidates under \(P^{(t)}\), and define the acquisition pool
\begin{equation}
\mathcal{A}_t
=
\bigcup_{d_j\in\mathcal{D}_t^\kappa}
\mathcal{V}_{d_j}^{L_1\cup L_2}
\setminus
\mathcal{V}_q^{(t)}.
\end{equation}
For each \(v\in\mathcal{A}_t\), SGUR scores
\begin{equation}
u_t(v)=\sum_{d_j\in\mathcal{D}_t^\kappa}\pi_v^{G_{d_j}},
\end{equation}
and selects the next query batch as
\begin{equation}
\mathcal{A}_t^\star
=
\mathrm{top}\text{-}b\big(\{u_t(v):v\in\mathcal{A}_t\}\big).
\end{equation}
The selected query is presented to the user, the returned observation is incorporated into \(\mathcal{O}_{t+1}\), and the pruning--alignment--ranking--acquisition loop repeats.

\paragraph{Termination and output.}
The interaction terminates when the turn budget \(T\) is exhausted or when no informative unobserved node remains. SGUR outputs the top-\(k\) candidate diagnoses under the terminal belief ranking:
\begin{equation}
\widehat{\mathcal{C}}(\mathcal{O}_\tau)
=
\mathrm{top}\text{-}k
\big(
\{P^{(\tau)}(d_j):G_{d_j}\in{G}_{\mathrm{cand}}^{(\tau)}\}
\big).
\end{equation}
For evaluation, we record the first round in which the ground-truth diagnosis enters the top-\(k\) set, yielding the MTTC metric used in Section~\ref{sec:exp-setup}.

\section{Experiments}
\label{sec:experiments}

\begin{table*}[t]
\centering
\caption{
Main accuracy, convergence, and efficiency results.
Final retrieval metrics are measured after round 3 in the complex two-symptom regime; convergence reports complex-regime Hit@3 across rounds; efficiency reports three-round average total tokens and latency.
All Hit@\(\!K\) values are percentages. SGUR achieves the best final ranking quality, lowest MTTC, and lowest inference cost.
}
\label{tab:main_unified}
\scriptsize
\setlength{\tabcolsep}{5.2pt}
\renewcommand{\arraystretch}{1.08}
\begin{tabular}{@{}lccccc ccc cc@{}}
\toprule
\multirow{2}{*}{Method}
& \multicolumn{5}{c}{Complex regime after R3}
& \multicolumn{3}{c}{Complex Hit@3}
& \multicolumn{2}{c}{Efficiency Avg.} \\
\cmidrule(lr){2-6} \cmidrule(lr){7-9} \cmidrule(lr){10-11}
& MTTC $\downarrow$
& MRR $\uparrow$
& Hit@1 $\uparrow$
& Hit@2 $\uparrow$
& Hit@3 $\uparrow$
& R1 & R2 & R3
& Tokens $\downarrow$
& Latency $\downarrow$ \\
\midrule
GraphRAG
& 1.74 & 0.71 & 61.20 & 77.60 & 83.10
& 59.60 & 76.90 & 83.10
& 10454.00 & 34.90 \\

NodeRAG
& 1.90 & 0.71 & 60.20 & 75.20 & 80.10
& 54.20 & 75.00 & 80.10
& 8461.10 & 15.20 \\

HippoRAG2
& 1.83 & 0.68 & 59.50 & 74.50 & 81.70
& 60.40 & 77.40 & 81.70
& 2352.80 & 11.40 \\

ArchRAG
& 2.01 & 0.65 & 56.70 & 70.40 & 76.10
& 56.70 & 70.40 & 76.10
& 2683.40 & 10.40 \\

\midrule
\textbf{SGUR (Ours)}
& \textbf{1.62} & \textbf{0.77} & \textbf{64.90} & \textbf{83.40} & \textbf{92.50}
& \textbf{61.40} & \textbf{84.70} & \textbf{92.50}
& \textbf{1845.40} & \textbf{8.90} \\
\bottomrule
\end{tabular}
\end{table*}

We evaluate SGUR on a multi-round diagnostic retrieval task that stresses inference under incomplete and progressively revealed evidence. The experiments are organized into two parts. Section~\ref{sec:exp-baseline} compares SGUR with graph-based RAG baselines in terms of accuracy, convergence, efficiency, and robustness to model scale. Section~\ref{sec:exp-ablation} isolates the contribution of SGUR's core mechanisms.

\paragraph{Experimental setup.}
\label{sec:exp-setup}
We construct a controlled multi-round diagnostic retrieval benchmark from \textbf{PrimeKG}~\citep{chandak2023building}. Each case is anchored by a ground-truth diagnosis \(d^\star\), and evidence is progressively revealed over three interaction rounds. PrimeKG provides typed symptom--anatomy--disease--treatment relations, enabling hierarchical graph construction and database-derived edge-prior initialization. By contrast, standard agent, tool-use, web-interaction, and GraphRAG benchmarks mainly evaluate final-answer accuracy, tool execution, short-horizon task completion, or retrieval-generation quality~\citep{liu2024agentbench,mialon2023gaia,qin2023toolllm,zhou2023webarena,xiao2025graphrag,xiang2025use}, but typically do not expose the typed hierarchy and relational statistics required to evaluate structure-grounded uncertainty reduction.

For each    disease, we randomly permute its associated symptoms once and reveal the same sequence to all methods to remove ordering randomness. We evaluate two regimes: \textbf{Simple Diagnosis}, with \(b=4\) newly revealed symptoms per round, and \textbf{Complex Diagnosis}, with \(b=2\) newly revealed symptoms per round. The simple regime contains 854 test cases and the complex regime contains 615 test cases, yielding 1,469 reasoning cases in total. We emphasize the complex regime in the main results because it best captures sparse and ambiguous evidence under partial observation. Additional construction details are provided in Appendix~\ref{app:data}.

\subsection{Comparison with Graph-RAG Baselines}
\label{sec:exp-baseline}

\textbf{Baselines and Multi-round Adaptation.} 
To verify the superiority of structure-grounded reasoning, we compare \textbf{SGUR} against four representative graph-RAG paradigms: the PageRank-based \textbf{HippoRAG2}~\citep{gutierrez2025rag}, node-centric \textbf{NodeRAG}~\citep{xu2025noderag}, community-level \textbf{GraphRAG}~\citep{microsoft2024graphrag}, and the hierarchical \textbf{ArchRAG}~\citep{wang2026archrag}. Since these baselines are primarily designed for single-shot inference, we adapt them to the three-round setting using a \textbf{textual context re-injection} scheme. In this setup, each baseline receives a compact textual summary of prior predictions and supporting evidence as conversational context in subsequent rounds. This ensures all methods access the same revealed evidence while preserving their native retrieval mechanisms. In contrast, SGUR facilitates a transition from context stacking to structured state evolution. By maintaining an incrementally updated query graph $G_q^{(t)}$, SGUR accumulates evidence directly as graph state, circumventing the semantic drift and computational redundancy inherent in repeatedly re-injecting dialogue history as text.

\textbf{Metrics, Results, and Implementation Details.} 
We assess retrieval effectiveness through \textbf{Hit@$K$} ($K \in \{1,2,3\}$), \textbf{MRR@R3}, and \textbf{Mean Turns to Confidence (MTTC)}, defined as the first round in which the ground-truth diagnosis enters the top-3 candidate set, with a penalty of $T+1=4$ for unretrieved cases. System efficiency is quantified via total token consumption and end-to-end latency. All models utilize \texttt{text-embedding-3-large}~\cite{openai2024embeddings} for indexing and retrieval. Unless otherwise specified, online inference is evaluated with \texttt{GPT-5.2}~\cite{openai2025gpt52}, and robustness to backbone scale is tested with \texttt{GPT-5.4-nano}~\cite{openai2026gpt54nano}. 

As detailed in Table~\ref{tab:main_unified}, SGUR achieves the strongest accuracy--efficiency profile in the complex diagnostic regime. It reaches \textbf{92.50\% Hit@3}, outperforming the strongest graph-RAG baseline by \textbf{9.40 percentage points}, while also obtaining the lowest MTTC. Its Hit@3 improves from 61.40\% to 92.50\% across rounds, indicating that the structured query state progressively narrows the candidate hypothesis space as new evidence is acquired. Furthermore, SGUR reduces token consumption by \textbf{5.70$\times$} and latency by \textbf{3.90$\times$} compared with GraphRAG. Together, these results show that SGUR improves diagnostic reliability and lowers the online cost of multi-turn inference by organizing reasoning through structure-grounded state updates, rather than relying solely on larger backbone models.

\textbf{Robustness to backbone scale.}
We further test whether SGUR's advantage persists when the backbone model is scaled down. Table~\ref{tab:scaling_compact} shows that SGUR with the smaller \textbf{GPT-5.4-nano} backbone remains above all graph-RAG baselines using the larger \textbf{GPT-5.2} backbone. Moreover, SGUR retains over \(99\%\) of its large-backbone Hit@3 after scaling down, while the baselines exhibit larger drops in ranking quality and convergence. This result supports the central claim that reliable multi-turn inference depends not only on backbone scale, but also on how the reasoning process is structurally organized. Full scaling results, including Hit@1 and Hit@2, are provided in Table~\ref{tab:scaling_combined_full} in Appendix~\ref{app:full-results1}.

\begin{table*}[tbp]
\centering
\caption{
Model-scaling stress test in the complex regime.
Each method is evaluated with large and small backbones; retention reports the small/large ratio.
SGUR preserves over \(99\%\) Hit@3 after backbone scaling and remains above all large-backbone graph-RAG baselines.
}
\label{tab:scaling_compact}
\small
\setlength{\tabcolsep}{8.0pt}
\renewcommand{\arraystretch}{1.12}
\begin{tabular}{@{}llrrr@{}}
\toprule
Method & Backbone & Hit@3 \(\uparrow\) & MRR \(\uparrow\) & MTTC \(\downarrow\) \\
\midrule
\multirow{3}{*}{GraphRAG}
& GPT-5.2 (L) & 83.10 & 0.71 & 1.74 \\
& GPT-5.4-nano (S) & 74.70 & 0.62 & 2.02 \\
& Retention (\%) & 90.30 & 87.10 & 86.10 \\
\midrule
\multirow{3}{*}{NodeRAG}
& GPT-5.2 (L) & 80.10 & 0.70 & 1.90 \\
& GPT-5.4-nano (S) & 73.60 & 0.60 & 2.12 \\
& Retention (\%) & 91.90 & 84.50 & 89.60 \\
\midrule
\multirow{3}{*}{HippoRAG2}
& GPT-5.2 (L) & 81.70 & 0.68 & 1.83 \\
& GPT-5.4-nano (S) & 79.80 & 0.60 & 2.11 \\
& Retention (\%) & 97.70 & 87.30 & 86.70 \\
\midrule
\multirow{3}{*}{ArchRAG}
& GPT-5.2 (L) & 76.10 & 0.65 & 2.01 \\
& GPT-5.4-nano (S) & 68.70 & 0.57 & 2.30 \\
& Retention (\%) & 90.20 & 87.40 & 87.30 \\
\midrule
\multirow{3}{*}{SGUR}
& GPT-5.2 (L) & \textbf{92.50} & \textbf{0.77} & \textbf{1.62} \\
& GPT-5.4-nano (S) & \textbf{92.40} & \textbf{0.73} & \textbf{1.63} \\
& Retention (\%) & \textbf{99.80} & \textbf{94.60} & \textbf{99.40} \\
\bottomrule
\end{tabular}
\end{table*}

\begin{figure}[t]
    \centering \includegraphics[width=0.8\linewidth,keepaspectratio]{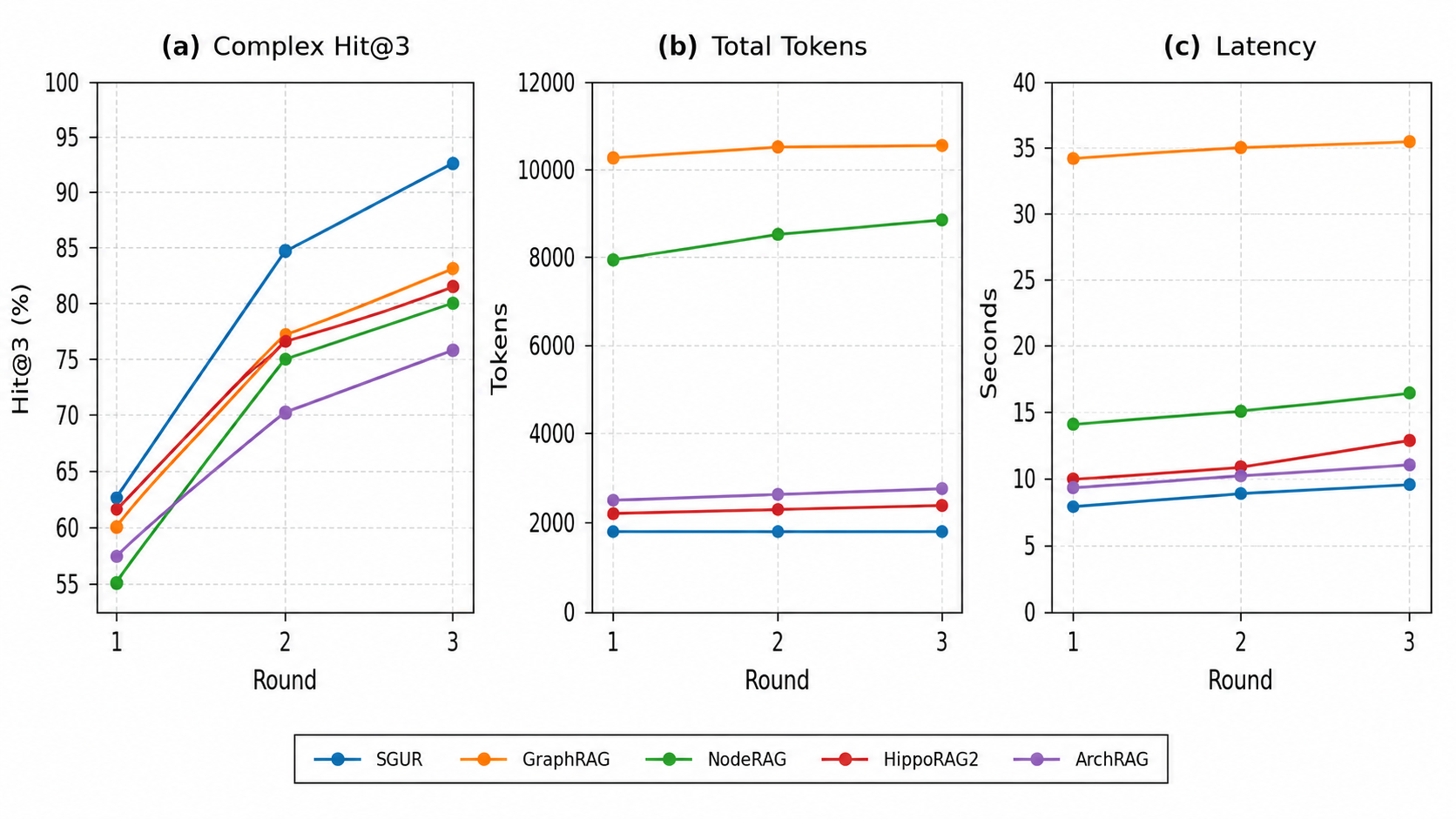}
    \caption{
    Accuracy--efficiency Pareto frontier in the complex regime.
    The x-axis reports average total tokens over three interaction rounds, and the y-axis reports final Hit@3 after round 3.
    Higher is better on the y-axis and lower is better on the x-axis.
    SGUR achieves the highest retrieval accuracy with the lowest inference cost.
    }
    \label{fig:pareto_tradeoff}
\end{figure}

\textbf{Accuracy--efficiency trade-off.}
Figure~\ref{fig:pareto_tradeoff} visualizes the system-level trade-off between final Hit@3 and average token cost. SGUR occupies the upper-left Pareto-optimal region, achieving the highest retrieval accuracy with the lowest inference budget among all methods. This demonstrates that SGUR shifts the accuracy--efficiency frontier, rather than improving accuracy by using larger prompts, longer interaction histories, or higher online computation.

\subsection{Ablation Study}
\label{sec:exp-ablation}

\textbf{Component contribution under sparse evidence.}
Table~\ref{tab:ablation_combined} jointly evaluates component necessity and evidence sufficiency. 
Removing the \(L_0\) latent routing layer causes the largest degradation, reducing final Hit@3 from \(92.50\%\) to \(68.90\%\) and increasing MTTC from \(1.62\) to \(2.31\). 
This shows that latent routing is not merely a computational shortcut: by aligning raw observations to database-level anchors before query-graph construction, it provides the grounding needed to build focused query graphs and retrieve relevant candidate hypotheses. 
Replacing partial transport with full FGW also causes a substantial drop, reducing Hit@3 to \(73.30\%\) and increasing MTTC to \(2.45\). 
This confirms that incomplete observations should be matched only to supported substructures rather than forced into full graph correspondence. 
Uniform importance reduces Hit@3 to \(83.40\%\), indicating that database-derived specificity priors help distinguish generic relations from diagnostically informative ones. 
Finally, semantic-only and structure-only variants remain competitive but consistently underperform the full model, showing that SGUR benefits from coupling node-level evidence with relational consistency.

\textbf{Effect of evidence sufficiency.}
The right block of Table~\ref{tab:ablation_combined} shows that SGUR's advantage is most pronounced when evidence is sparse. 
With only two observed symptoms, Full SGUR reaches \(61.40\%\) Hit@3, whereas removing latent routing drops performance to \(35.70\%\), and replacing partial transport with full FGW drops it further to \(15.30\%\). 
As additional observations are revealed, all variants improve, but the full model remains consistently strongest. 
This pattern supports the central mechanism of SGUR: structure-grounded routing, partial alignment, and specificity-aware weighting are most valuable when the agent must reduce uncertainty from incomplete evidence rather than simply rank hypotheses from abundant observations.

\begin{table*}[t]
\centering
\caption{
Combined ablation study in the complex regime.
The left block isolates each SGUR component at the final round, while the right block reports Hit@3 as observed evidence increases.
All Hit@3 and \(\Delta\)Hit@3 values are percentages.
}
\label{tab:ablation_combined}
\scriptsize
\setlength{\tabcolsep}{3.0pt}
\renewcommand{\arraystretch}{1.08}
\resizebox{\textwidth}{!}{
\begin{tabular}{@{}lccccccc@{}}
\toprule
\multirow{2}{*}{Variant}
& \multicolumn{4}{c}{Final complex-regime performance}
& \multicolumn{3}{c}{Evidence sufficiency: Hit@3} \\
\cmidrule(lr){2-5}
\cmidrule(lr){6-8}
& Hit@3 \(\uparrow\)
& \(\Delta\)Hit@3
& MRR \(\uparrow\)
& MTTC \(\downarrow\)
& 2 Obs.
& 4 Obs.
& 6 Obs. \\
\midrule
Full SGUR
& \textbf{92.50}
& --
& \textbf{0.77}
& \textbf{1.62}
& \textbf{61.40}
& \textbf{85.70}
& \textbf{92.50} \\

Semantic-only Alignment
& 91.90
& \(-0.60\)
& 0.75
& 1.64
& 59.00
& 85.30
& 91.90 \\

Structure-only Alignment
& 91.60
& \(-0.90\)
& 0.76
& 1.67
& 56.20
& 84.40
& 91.60 \\

Uniform Importance
& 83.40
& \(-9.10\)
& 0.64
& 2.01
& 38.20
& 71.80
& 83.40 \\

Full FGW
& 73.30
& \(-19.20\)
& 0.53
& 2.45
& 15.30
& 59.50
& 73.40 \\

w/o Latent Routing
& 68.90
& \(-23.60\)
& 0.57
& 2.31
& 35.70
& 59.80
& 68.90 \\
\bottomrule
\end{tabular}
}
\end{table*}

\section{Discussion and Conclusion}
\label{sec:discussion_conclusion}

Our results support a central thesis: reliable multi-turn inference depends not only on the scale of the backbone model, but also on how the reasoning process is organized. SGUR improves sequential inference by turning fragmented observations into a structured query state, aligning this state with candidate hypothesis graphs through prior-weighted partial matching, and actively selecting the next evidence node to reduce uncertainty. Across accuracy, convergence, latency, token cost, and model-scaling experiments, this structure-grounded design consistently improves performance under sparse and progressively revealed evidence. These findings suggest that reasoning frameworks can partially substitute for model scale by providing the structural discipline that free-form generation often lacks.

SGUR should nevertheless be interpreted as a research framework, not as a deployed clinical decision system. We instantiate the framework in sequential clinical diagnosis because it provides a demanding testbed with incomplete evidence, multi-round interaction, and structured hypothesis spaces. The system is not intended for real-world diagnosis, treatment recommendation, or medical decision-making without expert validation, clinician oversight, and appropriate regulatory review.

The framework also relies on several assumptions. It assumes that domain knowledge can be organized into a meaningful hierarchy, that observations can be mapped to standardized evidence nodes, and that the expert graph is sufficiently reliable during inference. Although PrimeKG provides a controlled large-scale benchmark, real-world settings may contain noisy observations, missing relations, evolving knowledge, and heterogeneous evidence sources. Future work should evaluate SGUR on broader long-horizon agent benchmarks and extend it to domains such as technical support, device maintenance, and software debugging, where agents must iteratively gather evidence, refine hypotheses, and make decisions under structural constraints.
\bibliographystyle{plainnat}
\bibliography{bib}

@inproceedings{liu2024agentbench,
  title         = {AgentBench: Evaluating LLMs as Agents},
  author        = {Liu, Xiao
                   and Yu, Hao
                   and Zhang, Hanchen
                   and Xu, Yifan
                   and Lei, Xuanyu
                   and Lai, Hanyu
                   and Gu, Yu
                   and Ding, Hangliang
                   and Men, Kaiwen
                   and Yang, Kejuan
                   and Zhang, Shudan
                   and Deng, Xiang
                   and Zeng, Aohan
                   and Du, Zhengxiao
                   and Zhang, Chenhui
                   and Shen, Sheng
                   and Zhang, Tianjun
                   and Su, Yu
                   and Sun, Huan
                   and Huang, Minlie
                   and Dong, Yuxiao
                   and Tang, Jie},
  booktitle     = {International Conference on Learning Representations},
  year          = {2024},
  url           = {https://arxiv.org/abs/2308.03688},
}

@article{mialon2023gaia,
  title         = {{GAIA}: A Benchmark for General {AI} Assistants},
  author        = {Mialon, Gr{\'e}goire and Fourrier, Cl{\'e}mentine and Swift, Craig and Wolf, Thomas and LeCun, Yann and Scialom, Thomas},
  journal       = {arXiv preprint arXiv:2311.12983},
  year          = {2023},
  url           = {https://arxiv.org/abs/2311.12983},
}

@article{qin2023toolllm,
  title         = {{ToolLLM}: Facilitating Large Language Models to Master 16000+ Real-World {API}s},
  author        = {Qin, Yujia
                   and Liang, Shihao
                   and Ye, Yining
                   and Zhu, Kunlun
                   and Yan, Lan
                   and Lu, Yaxi
                   and Lin, Yankai
                   and Cong, Xin
                   and Tang, Xiangru
                   and Qian, Bill
                   and Zhao, Sihan
                   and Hong, Lauren
                   and Tian, Runchu
                   and Xie, Ruobing
                   and Zhou, Jie
                   and Gerstein, Mark
                   and Li, Dahai
                   and Liu, Zhiyuan
                   and Sun, Maosong},
  journal       = {arXiv preprint arXiv:2307.16789},
  year          = {2023},
  url           = {https://arxiv.org/abs/2307.16789},
}

@article{edge2024local,
  title         = {From Local to Global: A Graph {RAG} Approach to Query-Focused Summarization},
  author        = {Edge, Darren
                   and Trinh, Ha
                   and Cheng, Newman
                   and Bradley, Joshua
                   and Chao, Alex
                   and Mody, Apurva
                   and Truitt, Steven
                   and Metropolitansky, Dasha
                   and Ness, Robert Osazuwa
                   and Larson, Jonathan},
  journal       = {arXiv preprint arXiv:2404.16130},
  year          = {2024},
  url           = {https://arxiv.org/abs/2404.16130},
}

@article{microsoft2024graphrag,
  title         = {From Local to Global: A Graph {RAG} Approach to Query-Focused Summarization},
  author        = {Edge, Darren
                   and Trinh, Ha
                   and Cheng, Newman
                   and Bradley, Joshua
                   and Chao, Alex
                   and Mody, Apurva
                   and Truitt, Steven
                   and Metropolitansky, Dasha
                   and Ness, Robert Osazuwa
                   and Larson, Jonathan},
  journal       = {arXiv preprint arXiv:2404.16130},
  year          = {2024},
  url           = {https://arxiv.org/abs/2404.16130},
}

@article{zhou2023webarena,
  title         = {{WebArena}: A Realistic Web Environment for Building Autonomous Agents},
  author        = {Zhou, Shuyan
                   and Xu, Frank F.
                   and Zhu, Hao
                   and Zhou, Xuhui
                   and Lo, Robert
                   and Sridhar, Abishek
                   and Cheng, Xianyi
                   and Ou, Tianyue
                   and Bisk, Yonatan
                   and Fried, Daniel
                   and Alon, Uri
                   and Neubig, Graham},
  journal       = {arXiv preprint arXiv:2307.13854},
  year          = {2023},
  url           = {https://arxiv.org/abs/2307.13854},
}

@article{xiao2025graphrag,
  title         = {{GraphRAG-Bench}: Challenging Domain-Specific Reasoning for Evaluating Graph Retrieval-Augmented Generation},
  author        = {Xiao, Yilin and Dong, Junnan and Zhou, Chuang and Dong, Su and Zhang, Qian-wen and Yin, Di and Sun, Xing and Huang, Xiao},
  journal       = {arXiv preprint arXiv:2506.02404},
  year          = {2025},
  url           = {https://arxiv.org/abs/2506.02404},
}

@article{xiang2025use,
  title         = {When to Use Graphs in {RAG}: A Comprehensive Analysis for Graph Retrieval-Augmented Generation},
  author        = {Xiang, Zhishang and Wu, Chuanjie and Zhang, Qinggang and Chen, Shengyuan and Hong, Zijin and Huang, Xiao and Su, Jinsong},
  journal       = {arXiv preprint arXiv:2506.05690},
  year          = {2025},
  url           = {https://arxiv.org/abs/2506.05690},
}

@misc{openai2025gpt52,
  title         = {Introducing {GPT-5.2}},
  author        = {{OpenAI}},
  year          = {2025},
  url           = {https://openai.com/index/introducing-gpt-5-2/},
  urldate       = {2026-04-16},
}

@misc{openai_gpt52_2025,
  title         = {Introducing {GPT-5.2}},
  author        = {{OpenAI}},
  year          = {2025},
  url           = {https://openai.com/index/introducing-gpt-5-2/},
  urldate       = {2026-04-16},
}

@misc{openai2026gpt54nano,
  title         = {Introducing {GPT-5.4} mini and nano},
  author        = {{OpenAI}},
  year          = {2026},
  month         = mar,
  url           = {https://openai.com/index/introducing-gpt-5-4-mini-and-nano/},
  urldate       = {2026-09-08},
}

@misc{openai2024embeddings,
  title         = {New Embedding Models and {API} Updates},
  author        = {{OpenAI}},
  year          = {2024},
  url           = {https://openai.com/index/new-embedding-models-and-api-updates/},
  urldate       = {2026-04-16},
  note          = {Introduces \texttt{text-embedding-3-large}},
}

@misc{openai_embeddings_2024,
  title         = {New Embedding Models and {API} Updates},
  author        = {{OpenAI}},
  year          = {2024},
  url           = {https://openai.com/index/new-embedding-models-and-api-updates/},
  urldate       = {2026-04-16},
  note          = {Introduces \texttt{text-embedding-3-large}},
}

@inproceedings{wang2026archrag,
  title         = {{ArchRAG}: Attributed Community-Based Hierarchical Retrieval-Augmented Generation},
  author        = {Wang, Shu and Fang, Yixiang and Zhou, Yingli and Liu, Xilin and Ma, Yuchi},
  booktitle     = {Proceedings of the AAAI Conference on Artificial Intelligence},
  volume        = {40},
  number        = {19},
  pages         = {15868--15876},
  year          = {2026},
  doi           = {10.1609/aaai.v40i19.38619},
  url           = {https://ojs.aaai.org/index.php/AAAI/article/view/38619},
}

@inproceedings{gorle2025quantifying,
  title         = {Quantifying Information Gain and Redundancy in Multi-Turn {LLM} Conversations},
  author        = {Gorle, Abhiram Rao and Yadav, Amit Kumar Singh and Weissman, Tsachy},
  booktitle     = {NeurIPS 2025 Workshop on Multi-Turn Interactions in Large Language Models},
  year          = {2025},
  url           = {https://openreview.net/forum?id=5gpABTkcUJ},
}

@article{laban2025llms,
  title         = {{LLM}s Get Lost in Multi-Turn Conversation},
  author        = {Laban, Philippe and Hayashi, Hiroaki and Zhou, Yingbo and Neville, Jennifer},
  journal       = {arXiv preprint arXiv:2505.06120},
  year          = {2025},
  url           = {https://arxiv.org/abs/2505.06120},
}

@article{scanlin2026representation,
  title         = {Representation Before Retrieval: Structured Patient Artifacts Reduce Hallucination in Clinical {AI} Systems},
  author        = {Scanlin, Joe and Cuesta, Armando and Varsavsky, Mart{\'\i}n},
  journal       = {medRxiv},
  year          = {2026},
  doi           = {10.64898/2026.02.13.26346256},
  url           = {https://www.medrxiv.org/content/10.64898/2026.02.13.26346256v1},
  note          = {Preprint, not peer reviewed},
}

@article{xu2025noderag,
  title         = {{NodeRAG}: Structuring Graph-Based {RAG} with Heterogeneous Nodes},
  author        = {Xu, Tianyang and Zheng, Haojie and Li, Chengze and Chen, Haoxiang and Liu, Yixin and Chen, Ruoxi and Sun, Lichao},
  journal       = {arXiv preprint arXiv:2504.11544},
  year          = {2025},
  doi           = {10.48550/arXiv.2504.11544},
  url           = {https://arxiv.org/abs/2504.11544},
}

@article{gutierrez2025rag,
  title         = {From {RAG} to Memory: Non-Parametric Continual Learning for Large Language Models},
  author        = {Jim{\'e}nez Guti{\'e}rrez, Bernal and Shu, Yiheng and Qi, Weijian and Zhou, Sizhe and Su, Yu},
  journal       = {arXiv preprint arXiv:2502.14802},
  year          = {2025},
  doi           = {10.48550/arXiv.2502.14802},
  url           = {https://arxiv.org/abs/2502.14802},
}

@article{lv2026all,
  title         = {All-Mem: Agentic Lifelong Memory via Dynamic Topology Evolution},
  author        = {Lv, Can and Chang, Heng and Tao, Shengyu and Chen, Mingju and Fan, Zhaoxin and Zhang, Ziwei and Guo, Yuchen and Zhou, Shiji},
  journal       = {arXiv preprint arXiv:2603.19595},
  year          = {2026},
  url           = {https://arxiv.org/abs/2603.19595},
}

@misc{bobrova2026clinical,
  title         = {Clinical Agentic Systems: Methodologies for Development, Orchestration, and Governance of Autonomous {AI} Agents in Healthcare Workflows---Alignment with 2026 {FDA/EMA} Guiding Principles},
  author        = {Bobrova, Natalia},
  year          = {2026},
  howpublished  = {SSRN},
  url           = {https://papers.ssrn.com/sol3/papers.cfm?abstract_id=6350259},
  note          = {Preprint, SSRN abstract 6350259},
}

@article{chung2025thinker,
  title         = {Thinker: Learning to Think Fast and Slow},
  author        = {Chung, Stephen and Du, Wenyu and Fu, Jie},
  journal       = {arXiv preprint arXiv:2505.21097},
  year          = {2025},
  url           = {https://arxiv.org/abs/2505.21097},
}

@article{bergamaschi2025fast,
  title         = {Fast, Slow, and Metacognitive Thinking in {AI}},
  author        = {Bergamaschi Ganapini, M
                   and Campbell, M
                   and Fabiano, F
                   and Horesh, L
                   and Lenchner, J
                   and Loreggia, A
                   and Mattei, N
                   and Rossi, F
                   and Srivastava, B
                   and Venable, K B},
  journal       = {npj Artificial Intelligence},
  volume        = {1},
  number        = {1},
  pages         = {27},
  year          = {2025},
  publisher     = {Nature Publishing Group UK London},
  doi           = {10.1038/s44387-025-00027-5},
  url           = {https://www.nature.com/articles/s44387-025-00027-5},
}

@article{you2026d,
  title         = {D-Mem: A Dual-Process Memory System for {LLM} Agents},
  author        = {You, Zhixing and Yuan, Jiachen and Cai, Jason},
  journal       = {arXiv preprint arXiv:2603.18631},
  year          = {2026},
  url           = {https://arxiv.org/abs/2603.18631},
}

@inproceedings{zhang2025leveraging,
  title         = {Leveraging Dual Process Theory in Language Agent Framework for Real-Time Simultaneous Human-{AI} Collaboration},
  author        = {Zhang, Shao
                   and Wang, Xihuai
                   and Zhang, Wenhao
                   and Li, Chaoran
                   and Song, Junru
                   and Li, Tingyu
                   and Qiu, Lin
                   and Cao, Xuezhi
                   and Cai, Xunliang
                   and Yao, Wen
                   and Zhang, Weinan
                   and Wang, Xinbing
                   and Wen, Ying},
  booktitle     = {Proceedings of the 63rd Annual Meeting of the Association for Computational Linguistics (Volume 1: Long Papers)},
  pages         = {4081--4108},
  year          = {2025},
  url           = {https://aclanthology.org/2025.acl-long.206/},
}

@article{zhang2025survey,
  title         = {A Survey of Graph Retrieval-Augmented Generation for Customized Large Language Models},
  author        = {Zhang, Qinggang
                   and Chen, Shengyuan
                   and Bei, Yuanchen
                   and Yuan, Zheng
                   and Zhou, Huachi
                   and Hong, Zijin
                   and Chen, Hao
                   and Xiao, Yilin
                   and Zhou, Chuang
                   and Dong, Junnan
                   and Chang, Yi
                   and Huang, Xiao},
  journal       = {arXiv preprint arXiv:2501.13958},
  year          = {2025},
  url           = {https://arxiv.org/abs/2501.13958},
}

@article{peng2025graph,
  title         = {Graph Retrieval-Augmented Generation: A Survey},
  author        = {Peng, Boci and Zhu, Yun and Liu, Yongchao and Bo, Xiaohe and Shi, Haizhou and Hong, Chuntao and Zhang, Yan and Tang, Siliang},
  journal       = {arXiv preprint arXiv:2408.08921},
  year          = {2024},
  eprint        = {2408.08921},
  archiveprefix = {arXiv},
  url           = {https://arxiv.org/abs/2408.08921},
}

@inproceedings{trivedi2023ircot,
  title         = {Interleaving Retrieval with Chain-of-Thought Reasoning for Knowledge-Intensive Multi-Step Questions},
  author        = {Trivedi, Harsh and Balasubramanian, Niranjan and Khot, Tushar and Sabharwal, Ashish},
  booktitle     = {Proceedings of the 61st Annual Meeting of the Association for Computational Linguistics (Volume 1: Long Papers)},
  pages         = {10014--10037},
  year          = {2023},
  url           = {https://aclanthology.org/2023.acl-long.557/},
}

@article{nori2025sequential,
  title         = {Sequential Diagnosis with Language Models},
  author        = {Nori, Harsha
                   and Daswani, Mayank
                   and Kelly, Christopher
                   and Lundberg, Scott
                   and Ribeiro, Marco Tulio
                   and Wilson, Marc
                   and Liu, Xiaoxuan
                   and Sounderajah, Viknesh
                   and Carlson, Jonathan
                   and Lungren, Matthew P.
                   and Gross, Bay
                   and Hames, Peter
                   and Suleyman, Mustafa
                   and King, Dominic
                   and Horvitz, Eric},
  journal       = {arXiv preprint arXiv:2506.22405},
  year          = {2025},
  url           = {https://arxiv.org/abs/2506.22405},
}

@inproceedings{xinjie2025reagent,
  title         = {{ReAgent}: Reversible Multi-Agent Reasoning for Knowledge-Enhanced Multi-Hop {QA}},
  author        = {Zhao, Xinjie
                   and Gao, Fan
                   and Song, Xingyu
                   and Chen, Yingjian
                   and Yang, Rui
                   and Fu, Yanran
                   and Wang, Yuyang
                   and Iwasawa, Yusuke
                   and Matsuo, Yutaka
                   and Li, Irene},
  booktitle     = {Proceedings of the 2025 Conference on Empirical Methods in Natural Language Processing},
  pages         = {4067--4089},
  year          = {2025},
  url           = {https://aclanthology.org/2025.emnlp-main.202.pdf},
}

@article{hu2026beyond,
  title         = {Beyond {RAG} for Agent Memory: Retrieval by Decoupling and Aggregation},
  author        = {Hu, Zhanghao and Zhu, Qinglin and Zhao, Runcong and Liang, Di and Yan, Hanqi and He, Yulan and Gui, Lin},
  journal       = {arXiv preprint arXiv:2602.02007},
  year          = {2026},
  url           = {https://arxiv.org/abs/2602.02007},
}

@article{li2026bayesrag,
  title         = {{BayesRAG}: Probabilistic Mutual Evidence Corroboration for Multimodal Retrieval-Augmented Generation},
  author        = {Li, Xuan and Wang, Yining and Luo, Haocai and Liu, Shengping and Liang, Jerry and Fu, Ying and {Weihuang} and Yu, Jun and Zhu, Junnan},
  journal       = {arXiv preprint arXiv:2601.07329},
  year          = {2026},
  url           = {https://arxiv.org/abs/2601.07329},
}

@article{wan2025mmgraphrag,
  title         = {{MMGraphRAG}: Bridging Vision and Language with Interpretable Multimodal Knowledge Graphs},
  author        = {Wan, Xueyao and Yu, Hang},
  journal       = {arXiv preprint arXiv:2507.20804},
  year          = {2025},
  url           = {https://arxiv.org/abs/2507.20804},
}

@article{yang2026autothinkrag,
  title         = {{AutoThinkRAG}: Complexity-Aware Control of Retrieval-Augmented Reasoning for Image-Text Interaction},
  author        = {Yang, Jiashu and Zhang, Chi and Wuerkaixi, Abudukelimu and Cheng, Xuxin and Liu, Cao and Zeng, Ke and Jia, Xu and Cai, Xunliang},
  journal       = {arXiv preprint arXiv:2603.05551},
  year          = {2026},
  url           = {https://arxiv.org/abs/2603.05551},
}

@misc{openai_gpt52_systemcard_2025,
  title         = {Update to {GPT-5} System Card: {GPT-5.2}},
  author        = {{OpenAI}},
  year          = {2025},
  url           = {https://openai.com/index/gpt-5-system-card-update-gpt-5-2/},
  urldate       = {2026-04-16},
}

@inproceedings{wei2022chain,
  title         = {Chain-of-Thought Prompting Elicits Reasoning in Large Language Models},
  author        = {Wei, Jason and Wang, Xuezhi and Schuurmans, Dale and Bosma, Maarten and Ichter, Brian and Xia, Fei and Chi, Ed H and Le, Quoc V and Zhou, Denny},
  booktitle     = {Advances in Neural Information Processing Systems},
  volume        = {35},
  pages         = {24824--24837},
  year          = {2022},
  url           = {https://proceedings.neurips.cc/paper_files/paper/2022/hash/9d5609613524ecf4f15af0f7b31abca4-Abstract-Conference.html},
}

@book{kahneman2011thinking,
  title         = {Thinking, Fast and Slow},
  author        = {Kahneman, Daniel},
  year          = {2011},
  publisher     = {Macmillan},
  url           = {https://us.macmillan.com/books/9780374275631/thinkingfastandslow/},
}

@inproceedings{yao2023tree,
  title         = {Tree of Thoughts: Deliberate Problem Solving with Large Language Models},
  author        = {Yao, Shunyu and Yu, Dian and Zhao, Jeffrey and Shafran, Izhak and Griffiths, Tom and Cao, Yuan and Narasimhan, Karthik},
  booktitle     = {Advances in Neural Information Processing Systems},
  volume        = {36},
  pages         = {11809--11822},
  year          = {2023},
  url           = {https://proceedings.neurips.cc/paper_files/paper/2023/hash/271db9922b8d1f4dd7aaef84ed5ac703-Abstract.html},
}

@inproceedings{besta2024graph,
  title         = {Graph of Thoughts: Solving Elaborate Problems with Large Language Models},
  author        = {Besta, Maciej
                   and Blach, Nils
                   and Kubicek, Ales
                   and Gerstenberger, Robert
                   and Podstawski, Michal
                   and Gianinazzi, Lukas
                   and Gajda, Joanna
                   and Lehmann, Tomasz
                   and Niewiadomski, Hubert
                   and Nyczyk, Piotr
                   and Hoefler, Torsten},
  booktitle     = {Proceedings of the AAAI Conference on Artificial Intelligence},
  volume        = {38},
  number        = {16},
  pages         = {17682--17690},
  year          = {2024},
  url           = {https://ojs.aaai.org/index.php/AAAI/article/view/29720},
}

@article{zhou2022least,
  title         = {Least-to-Most Prompting Enables Complex Reasoning in Large Language Models},
  author        = {Zhou, Denny
                   and Sch{\"a}rli, Nathanael
                   and Hou, Le
                   and Wei, Jason
                   and Scales, Nathan
                   and Wang, Xuezhi
                   and Schuurmans, Dale
                   and Cui, Claire
                   and Bousquet, Olivier
                   and Le, Quoc
                   and Chi, Ed},
  journal       = {arXiv preprint arXiv:2205.10625},
  year          = {2022},
  url           = {https://arxiv.org/abs/2205.10625},
}

@article{wang2022self,
  title         = {Self-Consistency Improves Chain of Thought Reasoning in Language Models},
  author        = {Wang, Xuezhi and Wei, Jason and Schuurmans, Dale and Le, Quoc and Chi, Ed and Narang, Sharan and Chowdhery, Aakanksha and Zhou, Denny},
  journal       = {arXiv preprint arXiv:2203.11171},
  year          = {2022},
  url           = {https://arxiv.org/abs/2203.11171},
}

@article{wang2024rat,
  title         = {{RAT}: Retrieval Augmented Thoughts Elicit Context-Aware Reasoning in Long-Horizon Generation},
  author        = {Wang, Zihao and Liu, Anji and Lin, Haowei and Li, Jiaqi and Ma, Xiaojian and Liang, Yitao},
  journal       = {arXiv preprint arXiv:2403.05313},
  year          = {2024},
  url           = {https://arxiv.org/abs/2403.05313},
}

@article{guo2024lightrag,
  title         = {{LightRAG}: Simple and Fast Retrieval-Augmented Generation},
  author        = {Guo, Zirui and Xia, Lianghao and Yu, Yanhua and Ao, Tu and Huang, Chao},
  journal       = {arXiv preprint arXiv:2410.05779},
  year          = {2024},
  url           = {https://arxiv.org/abs/2410.05779v1},
}

@article{chen2024multi,
  title         = {Knowledge Pyramid Construction for Multi-Level Retrieval-Augmented Generation},
  author        = {Chen, Rubing and Zhang, Xulu and Wu, Jiaxin and Fan, Wenqi and Wei, Xiao-Yong and Li, Qing},
  journal       = {arXiv preprint arXiv:2407.21276},
  year          = {2024},
  url           = {https://arxiv.org/abs/2407.21276},
  note          = {v1 (2024) was titled ``Multi-Level Querying using A Knowledge Pyramid''; retitled in v3 (2025)},
}

@article{nguyen2023linear,
  title         = {On a Linear Fused Gromov--Wasserstein Distance for Graph Structured Data},
  author        = {Nguyen, Dai Hai and Tsuda, Koji},
  journal       = {Pattern Recognition},
  volume        = {138},
  pages         = {109351},
  year          = {2023},
  publisher     = {Elsevier},
  url           = {https://www.sciencedirect.com/science/article/abs/pii/S0031320323000523},
}

@article{li2024structrag,
  title         = {{StructRAG}: Boosting Knowledge Intensive Reasoning of {LLM}s via Inference-Time Hybrid Information Structurization},
  author        = {Li, Zhuoqun and Chen, Xuanang and Yu, Haiyang and Lin, Hongyu and Lu, Yaojie and Tang, Qiaoyu and Huang, Fei and Han, Xianpei and Sun, Le and Li, Yongbin},
  journal       = {arXiv preprint arXiv:2410.08815},
  year          = {2024},
  url           = {https://arxiv.org/abs/2410.08815},
}

@inproceedings{hu2025grag,
  title         = {{GRAG}: Graph Retrieval-Augmented Generation},
  author        = {Hu, Yuntong and Lei, Zhihan and Zhang, Zheng and Pan, Bo and Ling, Chen and Zhao, Liang},
  booktitle     = {Findings of the Association for Computational Linguistics: NAACL 2025},
  pages         = {4145--4157},
  year          = {2025},
  url           = {https://aclanthology.org/2025.findings-naacl.232/},
}

@article{gutierrez2024hipporag,
  title         = {{HippoRAG}: Neurobiologically Inspired Long-Term Memory for Large Language Models},
  author        = {Jim{\'e}nez Guti{\'e}rrez, Bernal and Shu, Yiheng and Gu, Yu and Yasunaga, Michihiro and Su, Yu},
  journal       = {Advances in Neural Information Processing Systems},
  volume        = {37},
  pages         = {59532--59569},
  year          = {2024},
  url           = {https://arxiv.org/abs/2405.14831},
}

@inproceedings{zhu2025knowledge,
  title         = {Knowledge Graph-Guided Retrieval Augmented Generation},
  author        = {Zhu, Xiangrong and Xie, Yuexiang and Liu, Yi and Li, Yaliang and Hu, Wei},
  booktitle     = {Proceedings of the 2025 Conference of the Nations of the Americas Chapter of the Association for Computational Linguistics: Human Language Technologies (Volume 1: Long Papers)},
  pages         = {8912--8924},
  year          = {2025},
  url           = {https://aclanthology.org/2025.naacl-long.449/},
}

@article{han2025rag,
  title         = {{RAG} vs. {GraphRAG}: A Systematic Evaluation and Key Insights},
  author        = {Han, Haoyu
                   and Ma, Li
                   and Shomer, Harry
                   and Wang, Yu
                   and Lei, Yongjia
                   and Guo, Kai
                   and Hua, Zhigang
                   and Long, Bo
                   and Liu, Hui
                   and Aggarwal, Charu C.
                   and Tang, Jiliang},
  journal       = {arXiv preprint arXiv:2502.11371},
  year          = {2025},
  url           = {https://arxiv.org/abs/2502.11371v2},
}

@article{jiang2024reasoning,
  title         = {Reasoning-Enhanced Healthcare Predictions with Knowledge Graph Community Retrieval},
  author        = {Jiang, Pengcheng and Xiao, Cao and Jiang, Minhao and Bhatia, Parminder and Kass-Hout, Taha and Sun, Jimeng and Han, Jiawei},
  journal       = {arXiv preprint arXiv:2410.04585},
  year          = {2024},
  url           = {https://arxiv.org/abs/2410.04585},
}

@inproceedings{liang2025kag,
  title         = {{KAG}: Boosting {LLM}s in Professional Domains via Knowledge Augmented Generation},
  author        = {Liang, Lei
                   and Bo, Zhongpu
                   and Gui, Zhengke
                   and Zhu, Zhongshu
                   and Zhong, Ling
                   and Zhao, Peilong
                   and Sun, Mengshu
                   and Zhang, Zhiqiang
                   and Zhou, Jun
                   and Chen, Wenguang
                   and others},
  booktitle     = {Companion Proceedings of the ACM on Web Conference 2025},
  pages         = {334--343},
  year          = {2025},
}

@article{bai2025fused,
  title         = {Fused Partial Gromov--Wasserstein for Structured Objects},
  author        = {Bai, Yikun and Wang, Shuang and Tran, Huy and Du, Hengrong and Wang, Juexin and Kolouri, Soheil},
  journal       = {arXiv preprint arXiv:2502.09934},
  year          = {2025},
  url           = {https://arxiv.org/abs/2502.09934},
}

@inproceedings{yao2023react,
  title         = {{ReAct}: Synergizing Reasoning and Acting in Language Models},
  author        = {Yao, Shunyu and Zhao, Jeffrey and Yu, Dian and Du, Nan and Shafran, Izhak and Narasimhan, Karthik and Cao, Yuan},
  booktitle     = {The Eleventh International Conference on Learning Representations},
  year          = {2023},
  url           = {https://arxiv.org/abs/2210.03629},
}

@article{wong2026medclarify,
  title         = {{MedClarify}: An Information-Seeking {AI} Agent for Medical Diagnosis with Case-Specific Follow-Up Questions},
  author        = {Wong, Hui Min and Heesen, Philip and Janetzky, Pascal and Bendszus, Martin and Feuerriegel, Stefan},
  journal       = {arXiv preprint arXiv:2602.17308},
  year          = {2026},
  url           = {https://arxiv.org/abs/2602.17308},
}

@inproceedings{hu2024uot,
  title         = {Uncertainty of Thoughts: Uncertainty-Aware Planning Enhances Information Seeking in Large Language Models},
  author        = {Hu, Zhiyuan and Liu, Chumin and Feng, Xidong and Zhao, Yilun and Ng, See-Kiong and Luu, Anh Tuan and He, Junxian and Koh, Pang Wei and Hooi, Bryan},
  booktitle     = {Advances in Neural Information Processing Systems},
  volume        = {37},
  pages         = {24181--24215},
  year          = {2024},
  url           = {https://arxiv.org/abs/2402.03271},
}

@article{wang2024healthq,
  title         = {{HealthQ}: Unveiling Questioning Capabilities of {LLM} Chains in Healthcare Conversations},
  author        = {Wang, Ziyu and Li, Hao and Huang, Di and Rahmani, Amir M.},
  journal       = {arXiv preprint arXiv:2409.19487},
  year          = {2024},
  url           = {https://arxiv.org/abs/2409.19487v1},
}

@inproceedings{lewis2020retrieval,
  title         = {Retrieval-Augmented Generation for Knowledge-Intensive {NLP} Tasks},
  author        = {Lewis, Patrick
                   and Perez, Ethan
                   and Piktus, Aleksandra
                   and Petroni, Fabio
                   and Karpukhin, Vladimir
                   and Goyal, Naman
                   and K{\"u}ttler, Heinrich
                   and Lewis, Mike
                   and Yih, Wen-tau
                   and Rockt{\"a}schel, Tim
                   and Riedel, Sebastian
                   and Kiela, Douwe},
  booktitle     = {Advances in Neural Information Processing Systems},
  volume        = {33},
  pages         = {9459--9474},
  year          = {2020},
  url           = {https://proceedings.neurips.cc/paper/2020/hash/6b493230205f780e1bc26945df7481e5-Abstract.html},
}

@inproceedings{jiang2023active,
  title         = {Active Retrieval Augmented Generation},
  author        = {Jiang, Zhengbao and Xu, Frank F and Gao, Luyu and Sun, Zhiqing and Liu, Qian and Dwivedi-Yu, Jane and Yang, Yiming and Callan, Jamie and Neubig, Graham},
  booktitle     = {Proceedings of the 2023 Conference on Empirical Methods in Natural Language Processing},
  pages         = {7969--7992},
  year          = {2023},
  url           = {https://aclanthology.org/2023.emnlp-main.495/},
}

@inproceedings{borgeaud2022improving,
  title         = {Improving Language Models by Retrieving from Trillions of Tokens},
  author        = {Borgeaud, Sebastian
                   and Mensch, Arthur
                   and Hoffmann, Jordan
                   and Cai, Trevor
                   and Rutherford, Eliza
                   and Millican, Katie
                   and Van Den Driessche, George Bm
                   and Lespiau, Jean-Baptiste
                   and Damoc, Bogdan
                   and Clark, Aidan
                   and De Las Casas, Diego
                   and Guy, Aurelia
                   and Menick, Jacob
                   and Ring, Roman
                   and Hennigan, Tom
                   and Huang, Saffron
                   and Maggiore, Loren
                   and Jones, Chris
                   and Cassirer, Albin
                   and Brock, Andy
                   and Paganini, Michela
                   and Irving, Geoffrey
                   and Vinyals, Oriol
                   and Osindero, Simon
                   and Simonyan, Karen
                   and Rae, Jack
                   and Elsen, Erich
                   and Sifre, Laurent},
  booktitle     = {International Conference on Machine Learning},
  pages         = {2206--2240},
  year          = {2022},
  organization  = {PMLR},
  url           = {https://proceedings.mlr.press/v162/borgeaud22a.html},
}

@article{gao2023retrieval,
  title         = {Retrieval-Augmented Generation for Large Language Models: A Survey},
  author        = {Gao, Yunfan and Xiong, Yun and Gao, Xinyu and Jia, Kangxiang and Pan, Jinliu and Bi, Yuxi and Dai, Yi and Sun, Jiawei and Wang, Meng and Wang, Haofen},
  journal       = {arXiv preprint arXiv:2312.10997},
  year          = {2023},
  url           = {https://arxiv.org/abs/2312.10997},
}

@inproceedings{wu2025medical,
  title         = {Medical Graph {RAG}: Evidence-Based Medical Large Language Model via Graph Retrieval-Augmented Generation},
  author        = {Wu, Junde and Zhu, Jiayuan and Qi, Yunli and Chen, Jingkun and Xu, Min and Menolascina, Filippo and Jin, Yueming and Grau, Vicente},
  booktitle     = {Proceedings of the 63rd Annual Meeting of the Association for Computational Linguistics (Volume 1: Long Papers)},
  pages         = {28443--28467},
  year          = {2025},
  url           = {https://aclanthology.org/2025.acl-long.1381/},
}

@inproceedings{fastowski2024understanding,
  title         = {Understanding Knowledge Drift in {LLM}s through Misinformation},
  author        = {Fastowski, Alina and Kasneci, Gjergji},
  booktitle     = {International Workshop on Discovering Drift Phenomena in Evolving Landscapes},
  pages         = {74--85},
  year          = {2025},
  organization  = {Springer},
  doi           = {10.1007/978-3-031-82346-6_5},
  url           = {https://link.springer.com/chapter/10.1007/978-3-031-82346-6_5},
}

@inproceedings{zheng2025rethinking,
  title         = {Rethinking the Reliability of Multi-Agent Systems: A Perspective from Byzantine Fault Tolerance},
  author        = {Zheng, Lifan and Chen, Jiawei and Yin, Qinghong and Zhang, Jingyuan and Zeng, Xinyi and Tian, Yu},
  booktitle     = {Proceedings of the AAAI Conference on Artificial Intelligence},
  volume        = {40},
  number        = {41},
  pages         = {35012--35020},
  year          = {2026},
  url           = {https://ojs.aaai.org/index.php/AAAI/article/view/40806},
  note          = {Extended version: arXiv:2511.10400},
}

@article{luo2023reasoning,
  title         = {Reasoning on Graphs: Faithful and Interpretable Large Language Model Reasoning},
  author        = {Luo, Linhao and Li, Yuan-Fang and Haffari, Gholamreza and Pan, Shirui},
  journal       = {arXiv preprint arXiv:2310.01061},
  year          = {2023},
  url           = {https://arxiv.org/abs/2310.01061},
}

@article{sun2023think,
  title         = {Think-on-Graph: Deep and Responsible Reasoning of Large Language Model on Knowledge Graph},
  author        = {Sun, Jiashuo and Xu, Chengjin and Tang, Lumingyuan and Wang, Saizhuo and Lin, Chen and Gong, Yeyun and Ni, Lionel M and Shum, Heung-Yeung and Guo, Jian},
  journal       = {arXiv preprint arXiv:2307.07697},
  year          = {2023},
  url           = {https://arxiv.org/abs/2307.07697},
}

@article{vayer2020fused,
  title         = {Fused {G}romov--{W}asserstein Distance for Structured Objects},
  author        = {Vayer, Titouan and Chapel, Laetitia and Flamary, R{\'e}mi and Tavenard, Romain and Courty, Nicolas},
  journal       = {Algorithms},
  volume        = {13},
  number        = {9},
  pages         = {212},
  year          = {2020},
  publisher     = {MDPI},
  doi           = {10.3390/a13090212},
  url           = {https://www.mdpi.com/1999-4893/13/9/212},
}

@inproceedings{scetbon2022linear,
  title         = {Linear-Time {G}romov--{W}asserstein Distances using Low Rank Couplings and Costs},
  author        = {Scetbon, Meyer and Peyr{\'e}, Gabriel and Cuturi, Marco},
  booktitle     = {International Conference on Machine Learning},
  pages         = {19347--19365},
  year          = {2022},
  organization  = {PMLR},
  url           = {https://proceedings.mlr.press/v162/scetbon22b.html},
}

@article{memoli2011gromov,
  title         = {Gromov--Wasserstein Distances and the Metric Approach to Object Matching},
  author        = {M{\'e}moli, Facundo},
  journal       = {Foundations of Computational Mathematics},
  volume        = {11},
  pages         = {417--487},
  year          = {2011},
  url           = {https://www.researchwithrutgers.org/en/publications/gromov-wasserstein-distances-and-the-metric-approach-to-object-ma/},
}

@inproceedings{titouan2019optimal,
  title         = {Optimal Transport for Structured Data with Application on Graphs},
  author        = {Vayer, Titouan and Chapel, Laetitia and Flamary, R{\'e}mi and Tavenard, Romain and Courty, Nicolas},
  booktitle     = {International Conference on Machine Learning},
  pages         = {6275--6284},
  year          = {2019},
  organization  = {PMLR},
  url           = {https://proceedings.mlr.press/v97/titouan19a.html},
}

@article{vayer2018optimal,
  title         = {Optimal Transport for Structured Data with Application on Graphs},
  author        = {Vayer, Titouan and Chapel, Laetitia and Flamary, R{\'e}mi and Tavenard, Romain and Courty, Nicolas},
  journal       = {arXiv preprint arXiv:1805.09114},
  year          = {2018},
  url           = {https://arxiv.org/abs/1805.09114},
}

@article{chapel2020partial,
  title         = {Partial Optimal Transport with Applications on Positive-Unlabeled Learning},
  author        = {Chapel, Laetitia and Alaya, Mokhtar Z and Gasso, Gilles},
  journal       = {Advances in Neural Information Processing Systems},
  volume        = {33},
  pages         = {2903--2913},
  year          = {2020},
  url           = {https://proceedings.neurips.cc/paper_files/paper/2020/hash/1e6e25d952a0d639b676ee20d0519ee2-Abstract.html},
}

@article{chandak2023building,
  title         = {Building a Knowledge Graph to Enable Precision Medicine},
  author        = {Chandak, Payal and Huang, Kexin and Zitnik, Marinka},
  journal       = {Scientific Data},
  volume        = {10},
  number        = {1},
  pages         = {67},
  year          = {2023},
  publisher     = {Nature Publishing Group},
  doi           = {10.1038/s41597-023-01960-3},
  url           = {https://doi.org/10.1038/s41597-023-01960-3},
}

\clearpage
\appendix

\section{Additional Experimental Details and Results}
\label{app:reproducibility}

This section provides the additional implementation details needed to reproduce our main experimental results, including benchmark construction, test case sampling, hyperparameters, prompts, the compute environment, code and data release, and comprehensive results.
\subsection{Source Data, Schema Construction, and Test Case Sampling}
\label{app:data}

\paragraph{Source data.} We use \textbf{PrimeKG}~\cite{chandak2023building}
 as the source of clinical knowledge. From the
full graph we retain four node types corresponding to our hierarchical schema:
disease ($L_3$), phenotype ($L_1$), anatomy ($L_2$), and
drug ($L_4$). All other node types and their incident edges are
removed.

\paragraph{Phenotype--anatomy linkage.} PrimeKG does not directly link phenotypes
to the body parts where they appear. The closest available connection goes
through genes: a phenotype is linked to genes, and those genes are linked to
the organs where they are expressed. However, this only tells us where the
relevant genes happen to be active at the molecular level, not where the
phenotype is actually felt or observed in the clinic. For example, a phenotype
like headache may be associated with genes expressed throughout the body,
which does not mean the phenotype manifests in all of those locations. For this reason, we do not use this indirect path. Instead, we build the
phenotype--anatomy edges directly using a prompted LLM as a medical knowledge
source. For each phenotype $s \in L_1$, we ask the LLM to list the body parts
where $s$ typically appears in clinical practice, restricting its answers to
the anatomy vocabulary already present in PrimeKG. These (phenotype, anatomy)
pairs are then added to ${G}_{\mathrm{db}}$ as $L_1 \!\to\! L_2$
edges, completing the four-layer schema.

\paragraph{Test case sampling.}
We sample disjoint test sets for the simple ($b{=}4$) and complex ($b{=}2$) regimes. For each regime, we draw diseases uniformly at random from the disease pool. For each sampled disease, symptoms are revealed over three rounds according to the disease's associated symptom set. If a disease has fewer than $3b$ associated symptoms, symptoms are re-sampled from its symptom set in subsequent rounds to complete the required observation budget, while preserving diversity within each round whenever possible.
\subsection{Hyperparameters}
\label{app:hyperparameters}

Table~\ref{tab:hyperparameters} summarizes the main SGUR hyperparameters used across all experiments. Below we provide the additional implementation-level settings required for full reproduction.

\begin{table}[H]
\centering
\small
\caption{Hyperparameter settings used in all SGUR experiments unless stated otherwise.}
\label{tab:hyperparameters}
\begin{tabular}{llc}
\toprule
\textbf{Symbol} & \textbf{Description} & \textbf{Value} \\
\midrule
$T$ & Turn budget & 3 \\
$b$ & Per-turn question batch size (complex regime) & 2 \\
$b$ & Per-turn question batch size (simple regime) & 4 \\
$k$ & Top-$k$ return list size & 3 \\
$\kappa$ & Top-$\kappa$ candidate pool for evidence acquisition & 5 \\
$\tau_{\mathrm{sim}}$ & Entity-matching cosine-similarity threshold & 0.85 \\
$\alpha$ & PFGW feature--structure trade-off (Eq.~\ref{eq:pfgw}) & 0.5 \\
$c_{\mathrm{mis}}$ & PFGW feature-mismatch penalty & 1 \\
$\beta$ & Belief-distribution temperature (Eq.~\ref{eq:belief}) & 1.0 \\
$N_{L_0}$
 & Number of $L_0$ routing clusters & 256 \\
\bottomrule
\end{tabular}
\end{table}

\paragraph{PFGW solver.} The PFGW in Eq.~\eqref{eq:pfgw} is solved with the Frank--Wolfe algorithm from the POT library, using the squared-loss coupling $L(x,y)=|x-y|^2$ and a maximum of $2000$ outer iterations. The structural cost matrices $C_{d_j}$ and $C_q^{(t)}$ are computed as directed shortest-path distances under the specificity-aware edge weights of Eq.~\eqref{eq:edge_weight}, then jointly rescaled to $[0,1]$. The partial transport mass $m$ is set to the inherited overlap mass between query and candidate marginals, allowing unmatched evidence and unmatched expert structure to remain uncoupled. The trade-off $\alpha$ and feature-mismatch penalty $c_{\mathrm{mis}}$ follow Table~\ref{tab:hyperparameters}.

\paragraph{$L_0$ pre-filter.} The auxiliary routing layer $L_0$ is constructed offline over the $L_1$ observation embeddings produced by \textbf{text-embedding-3-large}, in two stages. We first reduce dimensionality with UMAP (\texttt{n\_neighbors}$=30$, controlling the local-vs-global trade-off; \texttt{n\_components}$=10$ target dimensions; cosine metric), then cluster the projected vectors with HDBSCAN (\texttt{min\_cluster\_size}$=15$, the minimum cluster size; \texttt{min\_samples}$=5$, controlling cluster density). At query time, the top $10$ clusters ranked by cosine similarity between the query embedding and each cluster's centroid are forwarded to the schema-constrained extractor as the candidate observation pool, which then drives the path-compatibility filter that yields ${G}_{\mathrm{cand}}^{(t)}$.

\paragraph{Entity matching.} Extracted entities are aligned to graph nodes by cosine similarity at threshold $\tau_{\mathrm{sim}}$, retaining the top $2$ observation ($L_1$) matches per extracted symptom and the top $1$ localization ($L_2$) match per observation.

\paragraph{Baseline hyperparameters.} We use the official open-source implementations and default hyperparameters of each baseline, with two exceptions made for fair comparison: (i) the same embedding model (\textbf{text-embedding-3-large}) is used across all methods, and (ii) the same indexing LLM (\textbf{GPT-5.2}) is used during corpus construction.
\subsection{Prompts}
\label{app:prompts}

We document the prompts used for each LLM-invoked component below. All prompts are released in full with our code. 
\paragraph{SGUR: schema-constrained extractor.} A single prompt asks the LLM to read the patient's message and pull out two things: the observations being described (\(L_1\)) and the localizations where they are felt (\(L_2\)). To keep the model from guessing, we also pass in a shortlist of likely observations produced by the \(L_0\) pre-filter and instruct the LLM to choose only from this list. The output must be returned as JSON with three fields: observations and localizations. 

\begin{tcolorbox}[
    enhanced,
    colback=gray!5,
    colframe=black!70,
    boxrule=0.5pt,
    arc=2pt,
    left=6pt, right=6pt, top=6pt, bottom=6pt,
    fonttitle=\bfseries,
    title={Entity recognition prompt},
    label={lst:ner_prompt},
    breakable
]
\small
\textbf{System message:}
\begin{quote}
You are a medical Named Entity Recognizer (NER).

\textbf{Task.} Extract symptoms mentioned or strongly implied by the user query, and list exactly 3 organs most specifically associated with each symptom.

\textbf{Matching.}
\begin{itemize}\itemsep0pt
    \item The user may describe symptoms directly, indirectly, or in lay language.
    \item You MUST semantically match to the closest symptom names from the Symptom List.
    \item Use meaning-based matching, not exact keywords.
\end{itemize}

\textbf{Hard rules.}
\begin{enumerate}\itemsep0pt
    \item Choose ONLY from the Symptom List (never invent symptoms).
    \item Return minimum 1 and AT MOST 3 symptoms total.
    \item For each symptom, list 1 of the most directly affected organs that are implicated by that symptom.
\end{enumerate}

\textbf{Symptom List (flat list):} \texttt{\{routed\_observation\_shortlist\}}

\textbf{Organ List:} \texttt{\{organ\_vocabulary\}}

\textbf{Output format (JSON only):}
\begin{verbatim}
{"symptom name from list": ["organ1", "organ2", "organ3"], ...}
\end{verbatim}
\end{quote}

\textbf{User message:} \texttt{\{query\}}
\end{tcolorbox}

\paragraph{Baselines: multi-round adaptation prompt.} As described in
Section~\ref{sec:exp-setup}, single-shot baselines are adapted to the
multi-round protocol by prompting them at each round to summarize their
top-ranked diseases together with the supporting observations and
localizations. The exact prompt template used for this adaptation is
shown in the box below.

\begin{tcolorbox}[
  colback=gray!5,
  colframe=black!70,
  title=\textbf{Multi-Round Adaptation Prompt for Single-Shot Baselines},
  fonttitle=\bfseries,
  boxrule=0.5pt,
  arc=2pt,
  left=6pt,right=6pt,top=4pt,bottom=4pt,
  breakable
]
\small
\textbf{\textsc{Role.}} You are a clinical reasoning assistant performing
differential disease identification based \emph{only} on retrieved medical
evidence.

\medskip
\textbf{\textsc{Task.}} The user's query describes symptoms experienced by
a patient. Using only the retrieved information:
\begin{enumerate}[leftmargin=*,topsep=2pt,itemsep=1pt]
  \item Identify which diseases are supported by the patient's symptoms.
  \item For each candidate disease, extract the supporting
        \emph{observations} (matched symptoms) and
        \emph{localizations} (affected organs or anatomical sites)
        from the retrieved context.
  \item Select the ten most likely diseases.
\end{enumerate}

\medskip
\textbf{\textsc{Rules.}}
\begin{itemize}[leftmargin=*,topsep=2pt,itemsep=1pt]
  \item Return exactly the top 10 diseases.
  \item Base reasoning solely on the retrieved context.
  \item Do not introduce diseases, observations, or localizations
        absent from the retrieved context.
  \item Do not fabricate symptoms, organs, or clinical facts.
  \item A disease is more likely if more of its phenotypes match the
        patient's symptoms and its associated organs are consistent
        with the described localizations.
  \item If observation or localization support is weak or ambiguous,
        still return the best 10 candidates with whatever evidence
        is available (use an empty list if none).
  \item Do not explain general medical knowledge or include disclaimers.
\end{itemize}

\medskip
\textbf{\textsc{Output.}} Return the following raw JSON only:
\begin{lstlisting}[basicstyle=\ttfamily\footnotesize,backgroundcolor=\color{gray!10},frame=none,xleftmargin=4pt]
{
  "top_10_diseases": [
    {
      "disease_name": "<name>",
      "supporting_observations": ["<symptom>", "<symptom>"],
      "supporting_localizations": ["<organ>", "<organ>"]
    },
    ...
  ]
}
\end{lstlisting}
\end{tcolorbox}

\subsection{Compute Environment}
\label{app:compute}

All experiments are run on a single machine with an 11th Gen Intel Core i7-11370H processor (4 cores, 8 threads) and 16\,GB of RAM. LLM inference is performed through the TensorBlock API gateway, which provides unified access to the evaluated backbones (GPT-5.2 and GPT-5.4-nano from OpenAI). Graph operations and PFGW solving are implemented in Python using \texttt{networkx} and \texttt{POT}. The end-to-end latency reported in Table~\ref{tab:app_efficiency_full} includes API round-trip time and is measured under the same network conditions for all methods to avoid bias.

\subsection{Code and Data Release}
\label{app:release}

To support reproducibility, we will release the source code, experimental scripts, processed data splits, and evaluation protocols upon acceptance of the paper. The released materials will include the implementation of SGUR, baseline configuration files, and instructions for reproducing the main results reported in the paper.

\subsection{Full Accuracy, Convergence, and Efficiency Results}
\label{app:full-results1}

Table~\ref{tab:app_main_and_convergence} expands the main results by reporting the full retrieval metrics and round-wise Hit@3 trajectories in both simple and complex regimes. This table complements the compact summary in the main text and shows that SGUR's advantage is not limited to the final round: in the complex two-symptom regime, SGUR obtains the best Hit@3 at every interaction round and reaches the correct diagnosis in fewer turns, as reflected by the lowest MTTC. In the simple regime, SGUR is slightly below HippoRAG2 at the first round but improves more rapidly as additional evidence is acquired, achieving the best R2 and R3 Hit@3. These trends support the claim that structure-grounded alignment and active acquisition convert newly revealed evidence into faster candidate-space contraction.
\begin{table*}[t]
\centering
\caption{
Full retrieval performance and multi-round convergence. 
Final metrics are measured after round 3 in the complex two-symptom regime, while the simple and complex columns report Hit@3 across interaction rounds. 
All Hit@\(\!K\) values are percentages. SGUR achieves the best final ranking quality and the lowest MTTC, and its advantage becomes more pronounced as evidence accumulates.
}
\label{tab:app_main_and_convergence}
\scriptsize
\setlength{\tabcolsep}{5.0pt}
\renewcommand{\arraystretch}{1.08}
\begin{tabular}{@{}lccccc ccc ccc@{}}
\toprule
\multirow{2}{*}{Method} 
& \multicolumn{5}{c}{Final complex-regime performance (R3)} 
& \multicolumn{3}{c}{Simple Hit@3} 
& \multicolumn{3}{c}{Complex Hit@3} \\
\cmidrule(lr){2-6} \cmidrule(lr){7-9} \cmidrule(lr){10-12}
& MTTC $\downarrow$ 
& MRR@R3 $\uparrow$ 
& Hit@1 $\uparrow$ 
& Hit@2 $\uparrow$ 
& Hit@3 $\uparrow$ 
& R1 & R2 & R3 
& R1 & R2 & R3 \\
\midrule
GraphRAG  
& 1.74 & 0.71 & 61.20 & 77.60 & 83.10 
& 64.20 & 81.30 & 84.50 
& 59.60 & 76.90 & 83.10 \\

NodeRAG
& 1.90 & 0.71 & 60.20 & 75.20 & 80.10 
& 63.00 & 79.00 & 86.00 
& 54.20 & 75.00 & 80.10 \\

HippoRAG2 
& 1.83 & 0.68 & 59.50 & 74.50 & 81.70 
& \textbf{68.30} & 82.90 & 87.70 
& 60.40 & 77.40 & 81.70 \\

ArchRAG
& 2.01 & 0.65 & 56.70 & 70.40 & 76.10 
& 58.20 & 74.10 & 80.90
& 56.70 & 70.40 & 76.10 \\

\midrule
\textbf{SGUR (Ours)}
& \textbf{1.62} & \textbf{0.77} & \textbf{64.90} & \textbf{83.40} & \textbf{92.50} 
& 67.10 & \textbf{86.70} & \textbf{93.40} 
& \textbf{61.40} & \textbf{84.70} & \textbf{92.50} \\
\bottomrule
\end{tabular}
\end{table*}

Table~\ref{tab:app_efficiency_full} provides the full per-round efficiency breakdown used to compute the average token and latency results reported in the main text. SGUR keeps input tokens nearly constant from R1 to R3 and produces substantially fewer output tokens than all baselines, indicating that the online loop relies on structured query-state updates rather than repeated free-form generation. In contrast, graph-RAG baselines accumulate larger input contexts as previous dialogue is re-injected. This explains why SGUR is both more accurate and more efficient in the main results.

\begin{table*}[t]
\centering
\caption{
Full per-round inference efficiency breakdown. 
We report input tokens, output tokens, total tokens, and end-to-end latency across three interaction rounds. 
SGUR keeps total token usage nearly flat from R1 to R3 and achieves the lowest latency in every round, confirming that structured query-state updates avoid the growing cost of re-injecting accumulated dialogue history.
}
\label{tab:app_efficiency_full}
\scriptsize
\setlength{\tabcolsep}{2.8pt}
\renewcommand{\arraystretch}{1.08}
\begin{tabular}{@{}lccc ccc ccc ccc@{}}
\toprule
\multirow{2}{*}{Method}
& \multicolumn{3}{c}{Input Tokens $\downarrow$}
& \multicolumn{3}{c}{Output Tokens $\downarrow$}
& \multicolumn{3}{c}{Total Tokens $\downarrow$}
& \multicolumn{3}{c}{Latency (s) $\downarrow$} \\
\cmidrule(lr){2-4} \cmidrule(lr){5-7} \cmidrule(lr){8-10} \cmidrule(lr){11-13}
& R1 & R2 & R3
& R1 & R2 & R3
& R1 & R2 & R3
& R1 & R2 & R3 \\
\midrule
GraphRAG
& 9886.00 & 10076.00 & 10080.00 
& 376.00 & 452.00 & 491.00 
& 10263.00 & 10529.00 & 10570.00 
& 34.10 & 35.00 & 35.50 \\

NodeRAG
& 7609.00 & 8118.40 & 8394.70 
& 337.40 & 427.90 & 496.00 
& 7946.40 & 8546.30 & 8890.70 
& 14.10 & 15.10 & 16.40 \\

HippoRAG2
& 2014.20 & 2058.90 & 2116.10 
& 249.40 & 293.50 & 326.20 
& 2263.70 & 2352.40 & 2442.40 
& 10.10 & 11.00 & 13.00 \\

ArchRAG
& 2223.40 & 2283.10 & 2384.40 
& 326.10 & 402.30 & 431.00 
& 2549.50 & 2685.40 & 2815.40 
& 9.50 & 10.40 & 11.20 \\

\midrule
\textbf{SGUR (Ours)}
& \textbf{1755.50} & \textbf{1759.70} & \textbf{1765.50} 
& \textbf{78.60} & \textbf{79.40} & \textbf{85.40} 
& \textbf{1838.00} & \textbf{1843.20} & \textbf{1854.90} 
& \textbf{8.10} & \textbf{9.10} & \textbf{9.70} \\
\bottomrule
\end{tabular}
\end{table*}

\begin{table*}[t]
\centering
\caption{
Full model-scaling stress test in the complex regime.
Each method is evaluated under a Large (L) and Small (S) backbone, with retention reported in the third row of each method.
Hit@\(k\) and retention values are percentages.
Retention is computed as Small/Large for Hit@\(k\) and MRR; for MTTC, where lower is better, retention is computed as \(\min(\mathrm{L},\mathrm{S})/\max(\mathrm{L},\mathrm{S})\).
All metrics are computed after round 3.
}
\label{tab:scaling_combined_full}
\small
\setlength{\tabcolsep}{4.0pt}
\renewcommand{\arraystretch}{1.12}
\begin{tabular}{@{}llrrrrr@{}}
\toprule
Method & Backbone & Hit@1 & Hit@2 & Hit@3 & MRR & MTTC \\
\midrule
\multirow{3}{*}{GraphRAG}
& GPT-5.2 (L) & 61.20 & 77.60 & 83.10 & 0.71 & 1.74 \\
& GPT-5.4-nano (S) & 52.10 & 64.40 & 74.70 & 0.62 & 2.02 \\
& Retention (\%) & 85.10 & 83.00 & 90.30 & 87.10 & 86.10 \\
\midrule
\multirow{3}{*}{NodeRAG}
& GPT-5.2 (L) & 60.20 & 75.20 & 80.10 & 0.70 & 1.90 \\
& GPT-5.4-nano (S) & 48.50 & 65.40 & 73.60 & 0.60 & 2.12 \\
& Retention (\%) & 80.60 & 87.00 & 91.90 & 84.50 & 89.60 \\
\midrule
\multirow{3}{*}{HippoRAG2}
& GPT-5.2 (L) & 59.50 & 74.50 & 81.70 & 0.68 & 1.83 \\
& GPT-5.4-nano (S) & 54.30 & 71.60 & 79.80 & 0.60 & 2.11 \\
& Retention (\%) & 91.40 & 96.10 & 97.70 & 87.30 & 86.70 \\
\midrule
\multirow{3}{*}{ArchRAG}
& GPT-5.2 (L) & 56.70 & 70.40 & 76.10 & 0.65 & 2.01 \\
& GPT-5.4-nano (S) & 49.80 & 63.80 & 68.70 & 0.57 & 2.30 \\
& Retention (\%) & 87.80 & 90.60 & 90.20 & 87.40 & 87.30 \\
\midrule
\multirow{3}{*}{SGUR}
& GPT-5.2 (L) & \textbf{64.90} & \textbf{83.40} & \textbf{92.50} & \textbf{0.77} & \textbf{1.62} \\
& GPT-5.4-nano (S) & \textbf{60.40} & \textbf{80.50} & \textbf{92.40} & \textbf{0.73} & \textbf{1.63} \\
& Retention (\%) & \textbf{93.10} & \textbf{96.50} & \textbf{99.80} & \textbf{94.60} & \textbf{99.40} \\
\bottomrule
\end{tabular}
\end{table*}

\end{document}